\documentclass[table]{hfstyle/hf}
\usepackage{microtype}
\usepackage{hyperref}
\usepackage{url}
\usepackage{booktabs}
\usepackage{graphicx}
\usepackage{lineno}
\usepackage{enumitem}
\usepackage{listings}
\usepackage{svg}

\definecolor{darkbluehf}{rgb}{0, 0, 0.5}
\hypersetup{colorlinks=true, citecolor=darkbluehf, linkcolor=darkbluehf, urlcolor=darkbluehf}

\usepackage{amssymb}
\usepackage{multirow}
\usepackage{bigdelim}
\usepackage{todonotes}
\usepackage{longtable}
\usepackage{tabularray}
\usepackage{wrapfig}
\usepackage[most]{tcolorbox}
\usepackage{url}
\usepackage{xspace}
\usepackage{svg}
\usepackage[absolute]{textpos}

\usepackage{amsmath}
\usepackage{siunitx}
\usepackage{arydshln}
\usepackage{enumitem}
\usepackage{soul}
\usepackage{adjustbox}
\usepackage{pifont}
\usepackage{caption}
\usepackage{makecell}
\usepackage{tabularx}
\usepackage{array}

\DeclareCaptionLabelFormat{subfig}{{\normalfont(\alph{subfigure})}}
\newcommand{\tableicon}[2]{\IfFileExists{#1}{\includegraphics[height=#2]{#1}}{}}
\usepackage{bold-extra}

\usepackage{pgf-pie}
\usepackage{tikz}

\usepackage[utf8]{inputenc}
\usepackage[T1]{fontenc}

\definecolor{maroon}{HTML}{F26035}
\definecolor{yellow}{HTML}{FDBC42}
\definecolor{lavender}{HTML}{734f96}
\definecolor{darkergrey}{HTML}{444444}
\definecolor{midgrey}{HTML}{e6eded}
\definecolor{neutralEight}{HTML}{343434}
\definecolor{neutralFive}{HTML}{838383}
\definecolor{neutralThree}{HTML}{bebebe}
\definecolor{neutralOne}{HTML}{dedede}
\definecolor{lightgrey}{HTML}{fafcfc}

\usepackage{tikz}

\definecolor{darkred}{RGB}{156, 39, 33}
\definecolor{darkblue}{RGB}{31, 90, 153}
\definecolor{forestgreen}{rgb}{0.13, 0.55, 0.13}

\setlist[itemize]{leftmargin=*,itemsep=0em,parsep=0.3em,topsep=0.3em}

\ifxetex\else
\DeclareUnicodeCharacter{2212}{\ensuremath{-}}
\fi

\usepackage{url}
\definecolor{lightyellow}{rgb}{1, 0.95, 0.85}
\definecolor{graphicbackground}{rgb}{0.9765,0.9451,0.9059}
\definecolor{codebackground}{rgb}{0.8314,0.949,0.9882}

\usepackage[table]{xcolor}

\usepackage{amsmath,amsfonts,bm}

\def\eqref#1{equation~\ref{#1}}
\def\1{\bm{1}}

\DeclareMathAlphabet{\mathsfit}{\encodingdefault}{\sfdefault}{m}{sl}
\SetMathAlphabet{\mathsfit}{bold}{\encodingdefault}{\sfdefault}{bx}{n}

\newcommand{\stanford}{\raisebox{.28em}{\hspace{.05em}\includegraphics[height=.75em]{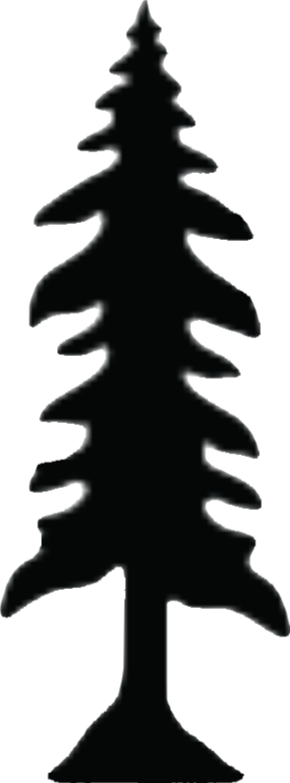}}\xspace}
\newcommand{\hf}{\raisebox{.28em}{\hspace{.05em}\includegraphics[height=.65em]{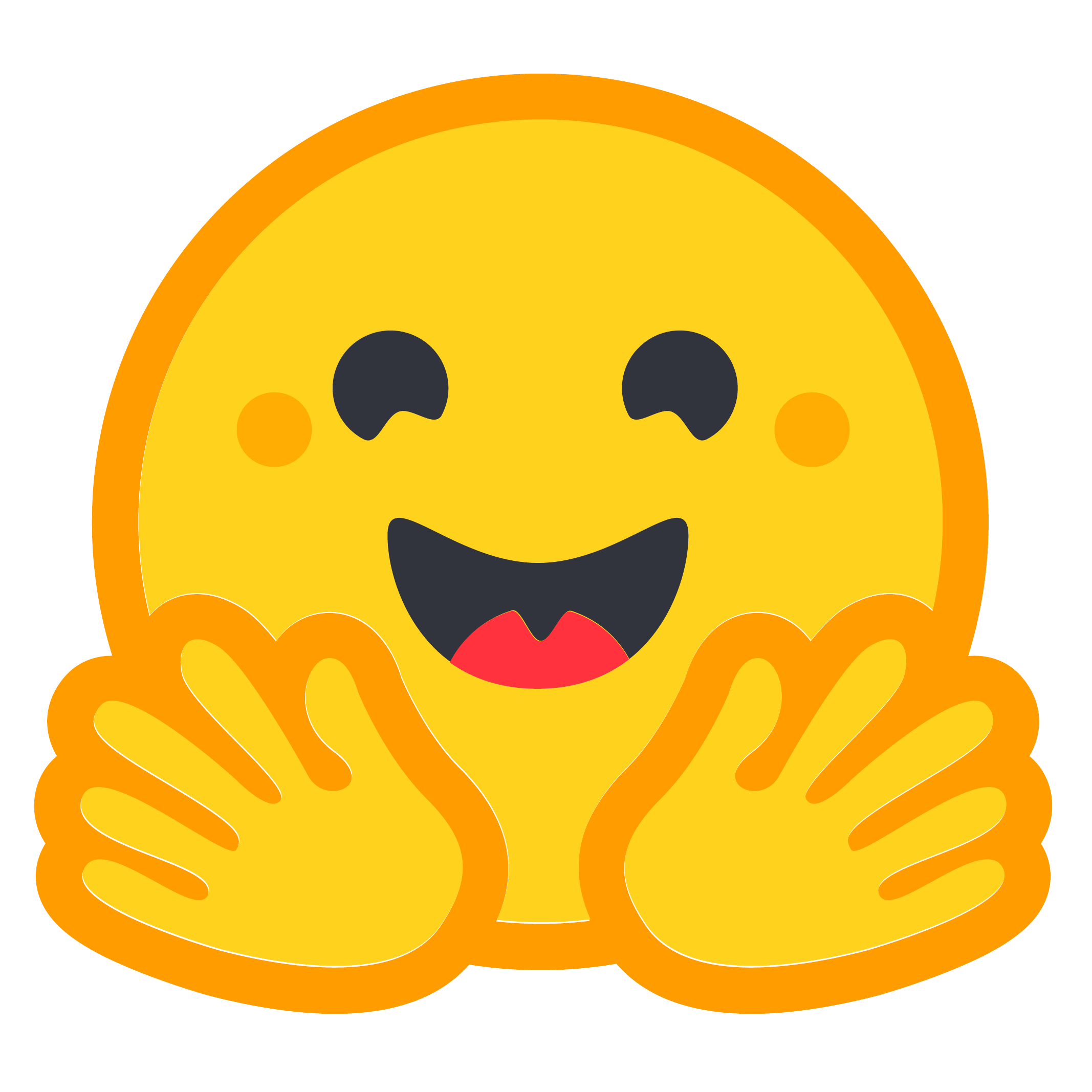}}\xspace}
\newcommand{\tum}{\raisebox{.28em}{\hspace{.05em}\includegraphics[height=.65em]{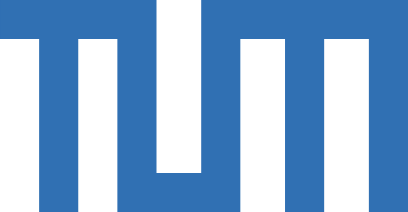}}\xspace}

\definecolor{lightgray}{gray}{0.9}

\title{FineVision: Open Data Is All You Need}

\authorOne[]{Luis Wiedmann$^{*}$\tum\hf}
\authorOne[]{Orr Zohar$^{*}$\stanford\hf}
\authorOne[]{Amir Mahla\hf}
\authorOne[]{Xiaohan Wang\stanford}
\authorOne[]{Rui Li\stanford}
\authorOne[]{Thibaud Frere\hf}
\authorOne[]{Leandro von Werra\hf}
\authorOne[]{Aritra Roy Gosthipaty\hf}
\authorOne[]{Andrés Marafioti\hf}

\newcommand{\huggingface}{\raisebox{-1.5pt}{\includegraphics[height=1.05em]{logos/hf.pdf}}\xspace}
\newcommand{\hfdataset}{\raisebox{-1.5pt}{\includegraphics[height=1.05em]{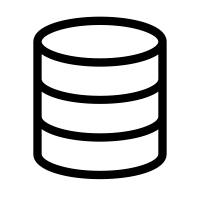}}\xspace}

\contribution[]{\hf Hugging Face, \tum Technical University of Munich, \stanford Stanford University~~\\$\textsuperscript{*}$Equal Contribution}

\setmainfigure{
\begin{figure}[!h]
    \centering
    \includegraphics[width=1\linewidth]{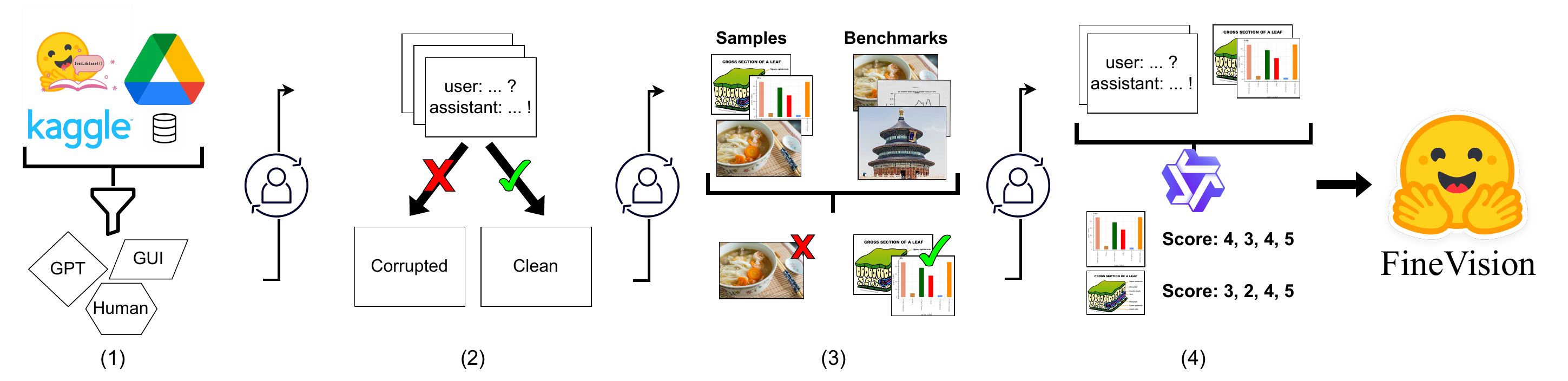}
    \caption{\textbf{Pipeline overview.} Left to right: (1) ingestion of raw sources; (2) canonicalization and image \& text cleaning; (3) 
    de-duplication and test-set decontamination using Self-Supervised Copy Detection embeddings~\citep{sscd2023}; (4) per-turn quality assessment with LLM/VLM-as-a-judge 
    \citep{zheng2023mtbench}. Each stage includes human checkpoints (mapping review, script sign-off, and post-conversion audits) to ensure faithful annotation
    consumption, consistent quality, and safety.}
    \label{fig:pipeline} 
\end{figure}
}

\abstract{The advancement of vision-language models (VLMs) is hampered by a fragmented landscape of inconsistent and contaminated public datasets.
We introduce \textit{FineVision}, a meticulously collected, curated, and unified corpus of 24 million samples; the largest open resource of its kind.
We unify more than 200 sources into 185 subsets via a semi-automated, human-in-the-loop pipeline: automation performs bulk ingestion and schema mapping, while reviewers audit mappings and spot-check outputs to verify faithful consumption of annotations, appropriate formatting and diversity, and safety; issues trigger targeted fixes and re-runs. The workflow further applies rigorous de-duplication within and across sources and decontamination against 66 public benchmarks.
FineVision also encompasses agentic/GUI tasks with a unified action space; reviewers validate schemas and inspect a sample of trajectories to confirm executable fidelity.
Models trained on FineVision consistently outperform those trained on existing open mixtures across a broad evaluation suite, underscoring the benefits of scale, data hygiene, and balanced automation with human oversight.
We release the corpus and curation tools to accelerate data-centric VLM research.}

\setmaintable{
\begin{table}[!h]
    \centering
    \begin{tabular}{l l @{\hspace{1.5cm}} l l}
        \hfdataset~{\sans \textbf{Dataset}} & \href{https://huggingface.co/datasets/HuggingFaceM4/FineVision}{\texttt{HuggingFaceM4/FineVision}} &
        
        \huggingface~{\sans \textbf{Blog Post}} & \href{https://huggingface.co/spaces/HuggingFaceM4/FineVision}{\texttt{HuggingFaceM4/FineVision}} \\

    \end{tabular}
\end{table}
}

\begin{document}

\maketitle

% Include FineVision sections
\section{Introduction}
\label{sec:intro}

The remarkable progress of vision-language models (VLMs) has been fueled by scaling both the capacity of the model and the training data.
However, the open research community faces a critical bottleneck: multimodal public datasets are fragmented, inconsistent, and often contaminated.
While proprietary models are trained on massive, curated corpora, open alternatives must stitch together many smaller, specialized datasets.
This misalignment not only hinders reproducibility, but also widens the performance gap between closed-source and open-source VLMs, limiting the community's ability to conduct robust, data-centric research.

Historically, early open aggregation efforts began with works like The Cauldron~\citep{laurencon2024matters}, followed by Cambrian-1~\citep{cambrian1} and LLaVA-OneVision~\citep{li2024llavaonevision}. These efforts were competitive at the time of release and laid the groundwork by unifying disparate sources.
Subsequently, leading open-source models have shifted toward much larger training mixtures that span hundreds of datasets and often combine open and closed sources; for example, Eagle2~\citep{eagle2} and PerceptionLM~\citep{cho2025perceptionlm} each report using on the order of 200 datasets.
As the field expands into agentic and GUI-grounded tasks, the need has moved from aggregation to principled, scalable curation.
The next frontier of VLM development requires datasets that are not only large-scale but also diverse and are engineered for emerging tasks.

To address this challenge, we introduce \textbf{FineVision}, a meticulously engineered corpus of over \textbf{24 million samples} with \textbf{17 million images}, \textbf{89 million turns}, and \textbf{9.5 billion answer tokens}. 
Our primary contribution is the collection, rigorous curation and open release of this dataset, providing a reliable, ready-to-use foundation for training and evaluating VLMs. 
To enable this, we developed a semi-automated, human-in-the-loop curation workflow that unifies over 200 sources and enforces a consistent chat schema. Automation performs bulk ingestion and schema mapping; reviewers then verify key steps through targeted audits and spot-checks. The pipeline conducts extensive cleaning -- removing corrupted images and malformed text, validating image-text alignment, and sanitizing unsafe content -- alongside rigorous de-duplication within and across sources and decontamination against 66 evaluation benchmarks to protect test integrity. Reviewers audited random samples to confirm faithful consumption of source annotations as well as appropriate formatting and diversity, and requested targeted fixes or re-runs when issues arose. For agentic/GUI data, they validated the unified action schema and inspected a small sample of trajectories to confirm executable fidelity. This review loop was repeated, as necessary, until quality criteria were met.

We validate FineVision through extensive experiments. Models trained on our corpus achieve state-of-the-art results among open-data VLMs, showing significant relative improvements over baselines: \textbf{27.7\%} over The Cauldron, \textbf{11.4\%} over Cambrian-7M, and \textbf{38.4\%} over LLaVA-OneVision on an average of 11 benchmarks. These results underscore the value of our principled approach to data hygiene and thoughtful integration.

We release FineVision and its associated resources to the public, aiming to democratize access to high-quality training data and catalyze the next wave of innovation in open VLM development.

\section{FineVision Curation}
\label{sec:data}

\textsc{FineVision} was created through a large-scale data curation effort to address the critical 
need for diverse, high-quality training data in the open-source VLM community. 
We unify over 200 public datasets through a semi-automated, human-in-the-loop process into a final corpus of 185 subsets. Automation performs bulk ingestion and schema mapping; reviewers then verify key steps for each dataset. Each source underwent a focused manual audit to accommodate its specific format and annotation style; for example, image QA, multi-image conversations, localized captions, and relational graphs.

We convert each dataset into a standardized chat format suitable for instruction tuning, using multiple conversational templates to ensure stylistic diversity and constructing multi-turn interactions where appropriate. Large language models were used to scale parts of the conversion; however, a reviewer remained in the loop to audit samples, confirm that source annotations were faithfully consumed, and request targeted fixes or re-runs when needed. This review loop was repeated, as necessary, until quality criteria were met.
This section details the curation workflow, from data sourcing (Sec.~\ref{sec:sources}) and schema unification (Sec.~\ref{sec:schema}) to cleaning and decontamination (Sec.~\ref{sec:cleaning} and~\ref{sec:dedup}), and the design choices that enabled this large-scale dataset.

\subsection{Data Sources}
\label{sec:sources}

Our data collection process aimed to be comprehensive, aggregating datasets from wherever they were publicly released by their original authors. 
We gathered over 200 datasets, sourcing data from a variety of locations: 
\begin{itemize}
    \item \textbf{Public Dataset Hubs:} Established platforms like Hugging Face Datasets, which host versioned and documented corpora.
    \item \textbf{Institutional and Cloud Storage:} Publicly shared links on institutional or personal cloud storage (\textit{e.g.}, Google Drive), a common hosting choice for academic releases.
    \item \textbf{Code Repositories:} GitHub repositories where datasets are shared alongside research code, often requiring custom extraction scripts.
    \item \textbf{Direct Web Downloads:} Project websites and other direct download links.
\end{itemize}
The full per-source breakdown is detailed in Appendix~\ref{app:datasets}.  After filtering and deduplication, we ended up with 185 subsets.

\subsection{From Heterogeneous Annotations to Unified Conversations}
\label{sec:schema}

Converting over 200 public datasets into a unified format suitable for instruction tuning was a significant engineering challenge. 
Each dataset arrived with its own annotation schema, task formulation, and data organization -- from simple image-caption pairs to complex multi-page document QA with derivations and spatial grounding. 
This subsection details our systematic approach to transforming this heterogeneous collection into high-quality conversational training data.

\paragraph{Semi-automated conversion pipeline.}
We developed a hybrid approach combining LLM assistance with human expertise. 
Using Claude as an agent, we broke down each dataset conversion into manageable subtasks: 
(1) deep annotation analysis to understand the structure and semantics of each dataset, 
(2) strategy design to map source annotations to conversational format, 
(3) script implementation with extensive validation, and 
(4) quality verification through sampling and then manual human inspection. 
This approach allowed us to maintain consistency across conversions while adapting to the unique requirements of each dataset. 
Every conversion script was manually reviewed and tested before full-scale processing.

\paragraph{Human-in-the-loop quality control.}
We prioritized automation while preserving accountability through targeted oversight. For each dataset, a reviewer (i) assessed the mapping plan and template choices, (ii) examined a dry-run of the converter, and (iii) audited a random sample of outputs to verify complete annotation consumption and appropriate formatting and diversity. When issues arose (\textit{e.g.}, missed fields or brittle templates), reviewers issued focused guidance and re-ran the affected stage. For agentic/GUI data, reviewers additionally validated the unified action schema and inspected a small sample of trajectories to confirm executable fidelity.

\paragraph{Unified conversational schema.}
All datasets converge to a standardized sample-level representation:
\[
\texttt{sample} = \{\texttt{images}, \texttt{texts}, \texttt{source}, \texttt{metadata}\},
\]
where \texttt{texts} contains a list of conversational turns alternating between \texttt{user} and \texttt{assistant} roles. 
Non-conversational sources (\textit{e.g.}, classification datasets) are transformed into natural QA pairs using carefully designed templates. 
We also experiment with converting single-turn QA datasets into multi-turn conversations by grouping multiple questions about the same image together, to create richer training signals that better leverage each image.
We preserve task-specific information in \texttt{metadata} for downstream filtering and analysis, including quality ratings, original sources, and confidence scores where available.

\newpage
\paragraph{Task-specific conversion strategies.}
We developed six core strategies to handle the diversity of supervision types while preserving their semantic richness. To ensure stylistic diversity, we randomized question templates and answer formats across conversions:
\begin{itemize}[leftmargin=1.5em, itemsep=2pt]
  \item \textbf{Visual QA}: Questions about the same image are grouped into multi-turn conversations. Multiple-choice questions include options in the prompt with answers providing both the selection and rationale. Question phrasings are varied (``What is...'', ``Can you identify...'', ``Tell me about...'') to avoid templatic patterns.
  \item \textbf{Captioning \& Description}: Ground-truth captions are wrapped with randomized instructional prompts (``Describe this image,'' ``What's shown here?'', ``Provide a detailed description of...'') to create natural QA pairs without altering the original descriptions.
  \item \textbf{Grounding \& Spatial Relations}: Spatial annotations (\textit{e.g.}, ``cat left of dog'') become yes/no questions with varied phrasings and explanatory answers. Bounding box coordinates are converted into natural language descriptions of spatial relationships (\textit{e.g.}, left, right, above), while the raw coordinates are normalized to a \texttt{(cx, cy, w, h)} format and preserved in metadata.
  \item \textbf{Document Understanding}: Multi-page documents are processed as image lists with questions threaded into conversations. Answers are enriched with available annotations like derivation steps, supporting facts, and answer types (arithmetic, extractive, etc.).
  \item \textbf{OCR \& Transcription}: We generate both exact \textit{transcription turns} and optional \textit{understanding turns} that explain the content's structure or meaning, particularly useful for mathematical expressions and handwritten text.
  \item \textbf{Classification \& Detection}: Binary or categorical labels are converted to decision questions with explanatory answers when auxiliary descriptions are available, maintaining the educational value of the original annotations.
\end{itemize}

\paragraph{Action-space unification for GUI data.}
To enable novel capabilities in agentic vision tasks, we include multiple GUI automation datasets where a major challenge is the lack of standardization in action spaces: different sources define heterogeneous function signatures, parameter naming conventions, and action taxonomies.
To address this, we built a data transformation pipeline on top of the open-source datasets used in~\citet{xu_aguvis_2025}.
Our pipeline includes (i) a parser that extracts and normalizes arbitrary function signatures, ensuring consistent parameter ordering and reconstruction, and (ii) an action conversion module that maps all action representations into a unified schema.
This process enforces consistent function and parameter naming, and produces a coherent, typed action schema.
Screen coordinates are expressed in normalized form [0,1] to ensure resolution-agnostic training.
By unifying the action space, we enable cross-domain training and allow models to learn coherent action patterns across heterogeneous GUI environments (desktop, mobile, or browser).
See Appendix~\ref{app:action_space} for further details.

\subsection{Cleaning}
\label{sec:cleaning}
Our workflow includes several automated cleaning and validation steps to handle edge cases common in large-scale data aggregation, such as corrupted files, malformed annotations, and inconsistent text formatting.

\paragraph{Images.}
We perform automatic image validation, decoding with robust backends, to discard undecodable, corrupted, or zero-byte images. We also orient images via EXIF metadata and convert all formats to RGB. Any samples with failed image I/O are dropped from the dataset and we cap the image size at 2048px for the longest side while preserving the aspect ratio.

\paragraph{Text.}
Text content is normalized to enforce UTF-8 encoding, strip control characters, standardize punctuation and quotes, and remove artifacts such as base64 blobs. We collapse repeated tokens (\textit{e.g.}, \texttt{!!!!} \(\rightarrow\) \texttt{!}) and remove turns with empty or degenerate answers (\textit{e.g.}, single-character repeats). To filter outliers and ensure training stability, every turn is capped at a combined question and answer length of 8192 tokens.

\begin{figure}[ht]
  \centering
  \includegraphics[width=\linewidth]{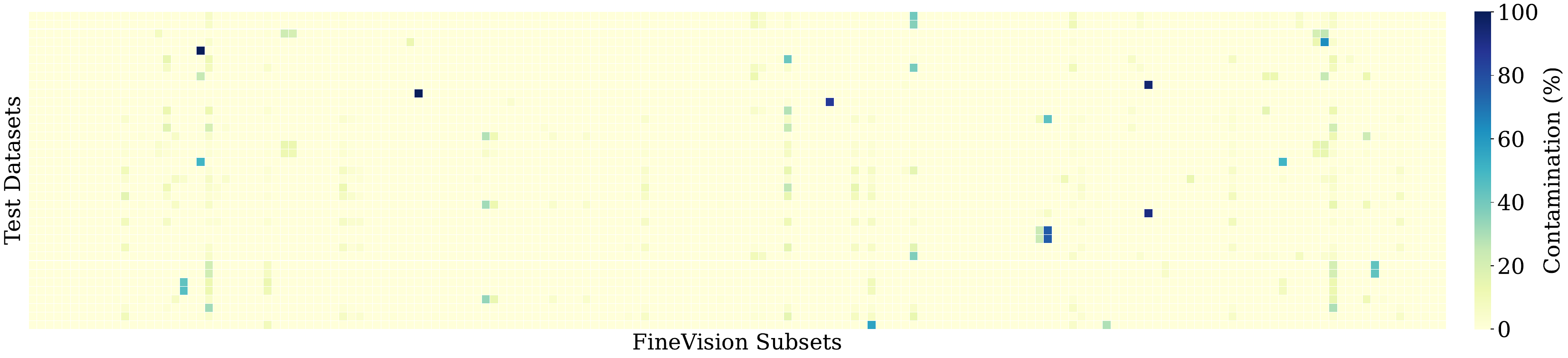}
  \includegraphics[width=\linewidth]{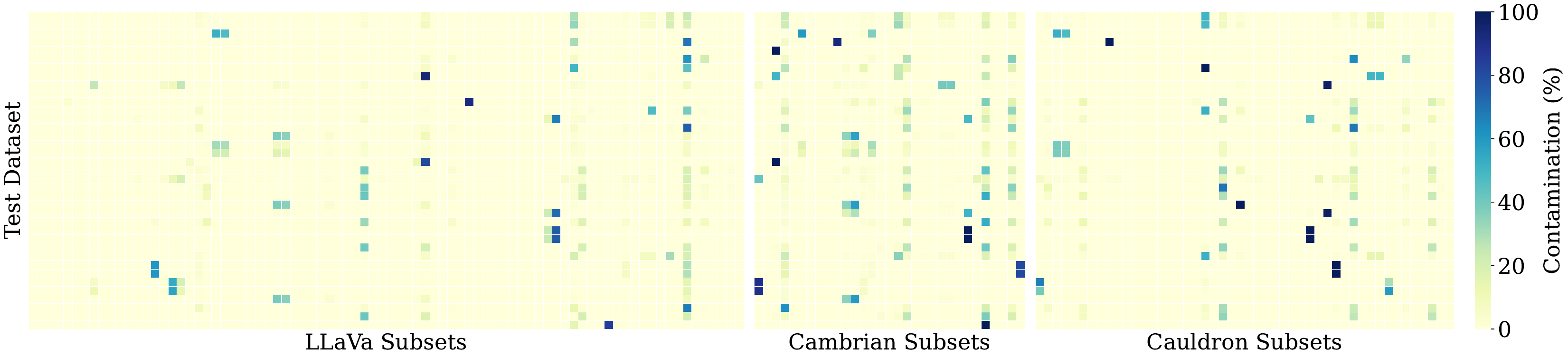}
  \caption{\textbf{Decontamination report.}
Per-benchmark contamination heatmap for FineVision and comparable open-source alternatives (rows: benchmarks, columns: datasets subsets). FineVision's contamination is sparse and concentrated in a few subsets and benchmarks, and consistently lower than the baselines.}
  \label{fig:decontam}
\end{figure}

\subsection{Near-Duplicate and Contamination Control}
\label{sec:dedup}

We perform hygiene in two stages using self-supervised copy-detection descriptors (SSCD) \citep{sscd2023} and cosine similarity:

\begin{enumerate}[leftmargin=1.5em, itemsep=2pt]
  \item \textbf{Intra-Dataset:} cluster visually near-identical images within FineVision, to merge samples from the same image into a multi-turn conversation or a merged subset.
  \item \textbf{Test-Set Decontamination:} identifying training images similar to evaluation images from 66 public VLM benchmarks (via embeddings computed once from the same SSCD model), mitigating train-test leakage \citep{razeghi2022impact}, see Fig. \ref{fig:decontam}.
\end{enumerate}

All stages share a threshold \( \tau = 0.95 \) on cosine similarity, erring on the conservative side to reduce false positives (examples of the different scenarios are in the Appendix, Fig. \ref{fig:dup_examples}).

\paragraph{Intra-dataset duplicates.}
We flag subsets for potential overlap with each other using the SSCD+cosine pipeline and manually inspect them before potentially merging into a single subset (e.g., we merged three commonly found online variants of \texttt{ai2d} into a single \texttt{ai2d\_merged} subset). We additionally experiment with generally merging multiple individual questions for the same image into a multi-turn conversation, but this did not result in improved performance during our ablations.

\paragraph{Contamination measurement against public benchmarks.}

Following the same SSCD+cosine protocol, we embed all images from 66 test sets included in \texttt{lmms-eval} \citep{zhang2024lmms} and compute their max similarity to each training image.
Images with similarity \(\ge \tau\) are flagged, and we study the impact of removing them from training, but since this is not a definitive indicator of a contaminated \emph{sample}, we release FineVision in its original form.
Detailed description and statistics of contamination and performance drop across datasets are in the Appendix (Table \ref{tab:contamination_performance}) and contamination is visualized in Fig. \ref{fig:decontam}.
We release both the de-duplication pipeline\footnote{\url{https://github.com/huggingface/large-scale-image-deduplication}} and the precomputed SSCD embeddings for the public benchmarks\footnote{\url{https://huggingface.co/datasets/HuggingFaceM4/lmms-eval-embeddings}}.
\section{Exploring FineVision}
\label{sec:characterization}
We characterize FineVision along three key axes: category composition, turn quality, and visual diversity. 

\subsection{Category Composition}
\label{sec:char-category}
\begin{figure}[t]
  \centering
  \includegraphics[width=\linewidth]{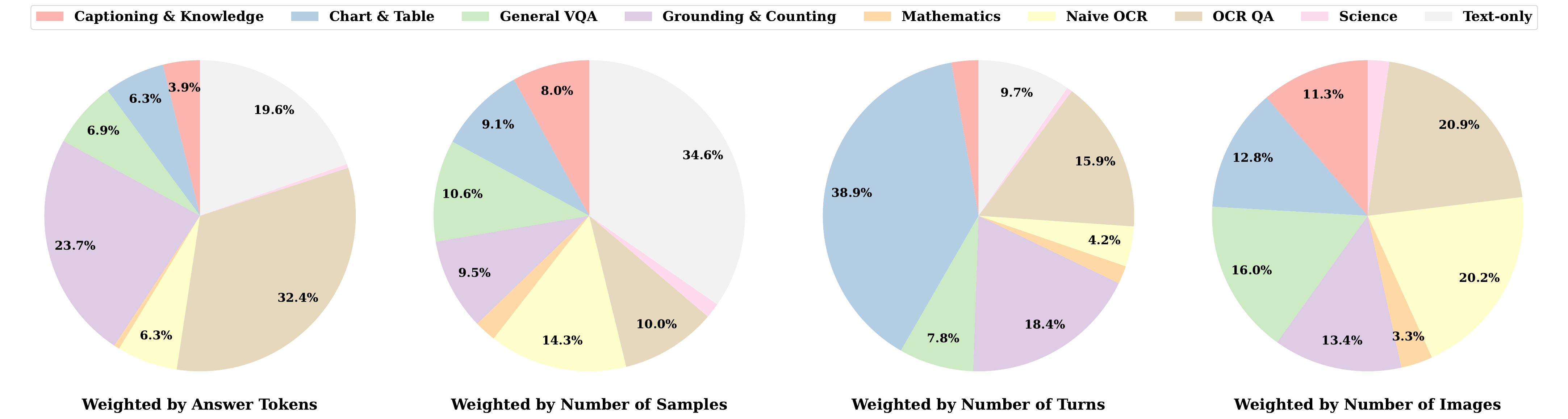}\vspace{-0.07in}
  \caption{\textbf{Category distribution.} Share of samples across the nine categories.
FineVision provides a good baseline mixture, which could be further tuned via up- and downsampling and in correlation with the quality ratings.}
  \label{fig:category_distribution}
\end{figure}

We categorize every FineVision subset into nine distinct categories, following \cite{eagle2}: \texttt{Captioning \& Knowledge},  \texttt{Chart \& Table},  \texttt{General VQA},  \texttt{Grounding \& Counting},  \texttt{Mathematics},  \texttt{Naive OCR}, \texttt{OCR QA}, \texttt{Science} and \texttt{Text-only}.
We analyze the resulting category composition along multiple axes: number of images, samples, turns, and answer tokens (see Fig.~\ref{fig:category_distribution}).
Samples from  \texttt{Chart \& Table} usually lend themselves well to multi-turn conversations, since multiple similar questions can be asked for a single chart.
Samples from  \texttt{OCR QA} tend to have longer answers, since they aim at detailed document understanding, which are rarely answered with a short sentence. For in-depth statistics on token length, conversation turns, and image resolution by category, see Appendix~\ref{sec:char-len-res}.

\begin{figure}[p]
    \centering
    \begin{subfigure}[b]{0.48\linewidth}
      \centering
      \includegraphics[width=\linewidth]{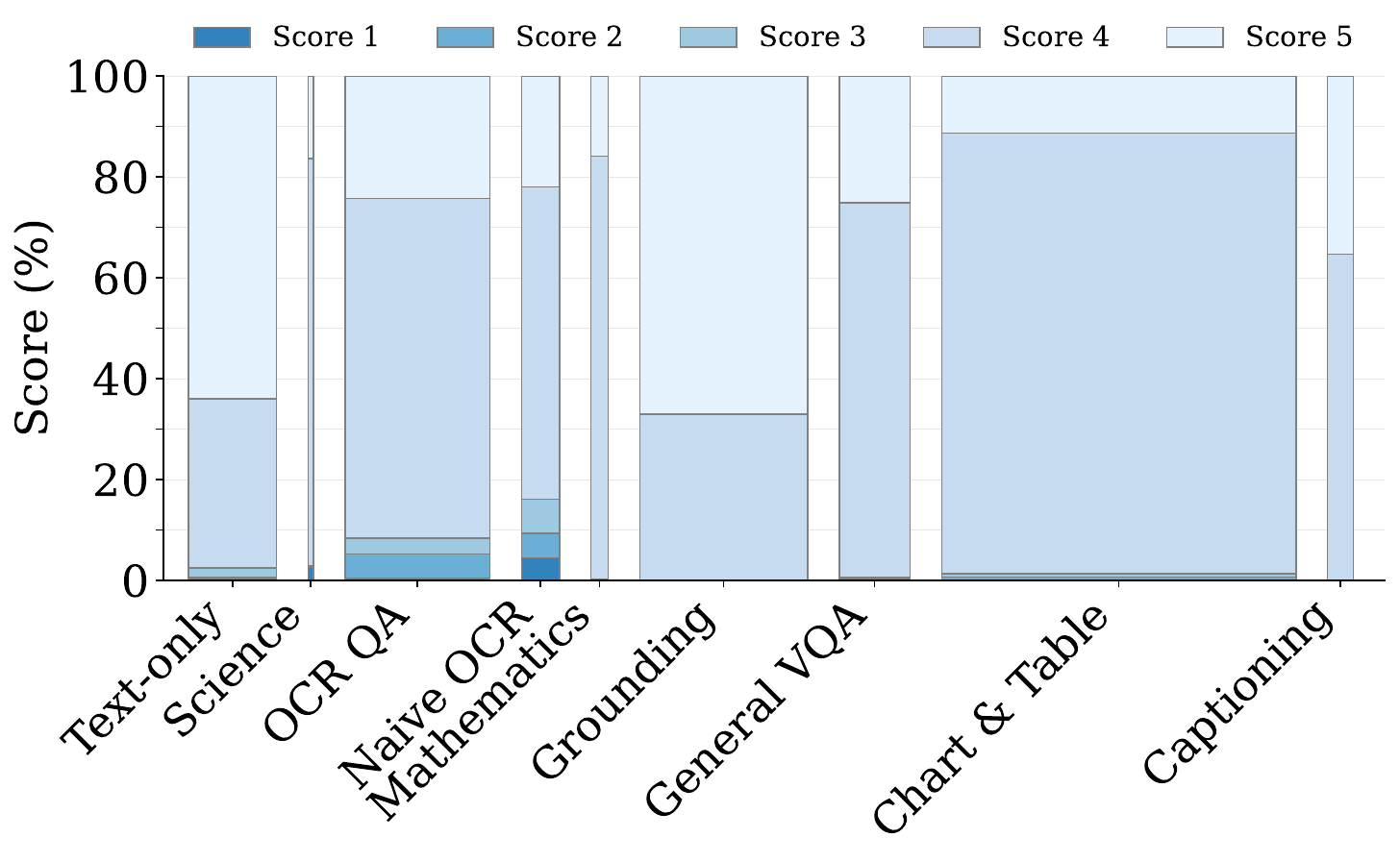}
      \caption{Formatting}
      \label{fig:formatting_scores}
    \end{subfigure}
    \hfill
    \begin{subfigure}[b]{0.48\linewidth}
      \centering
      \includegraphics[width=\linewidth]{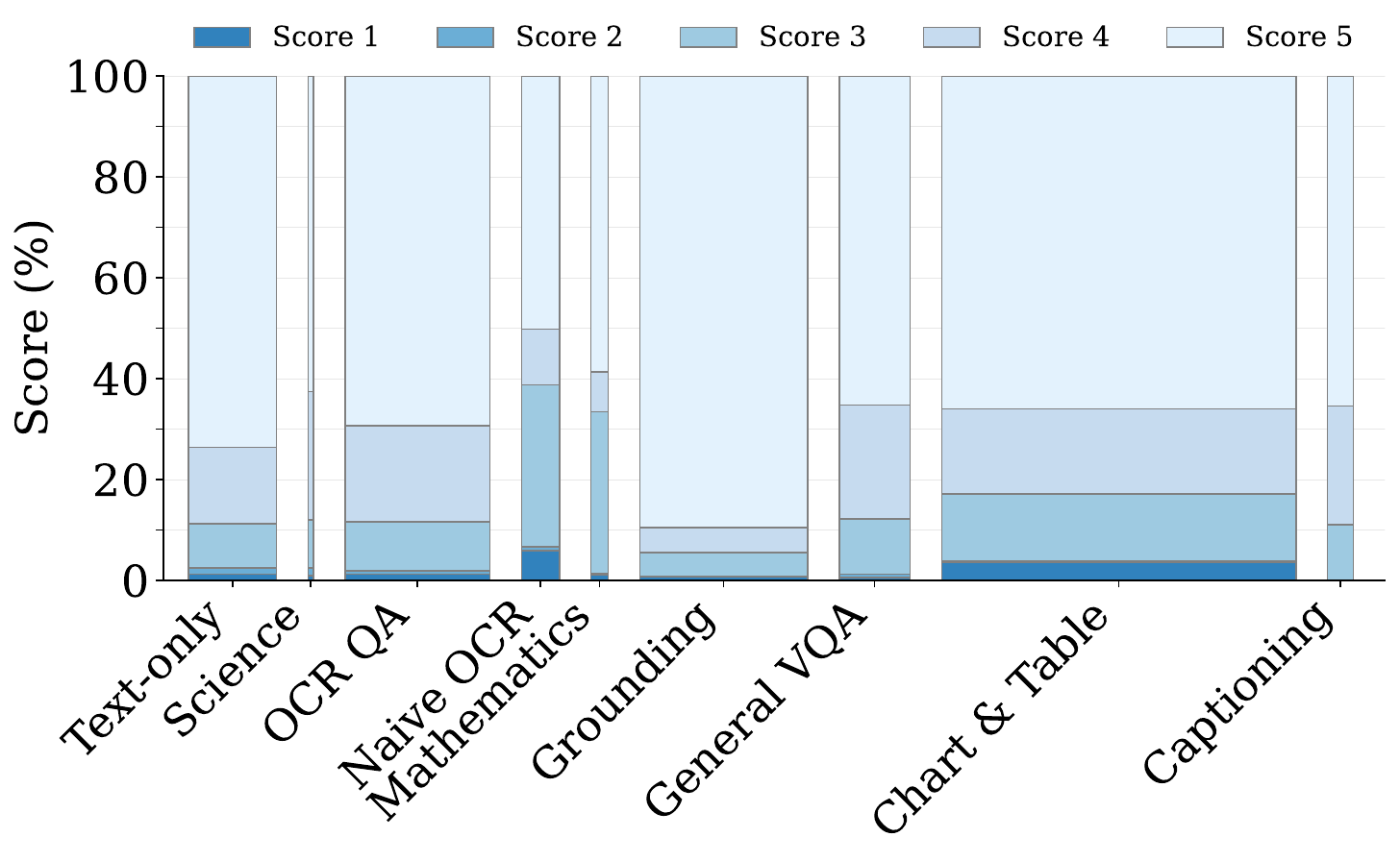}
      \caption{Relevance}
      \label{fig:relevance_scores}
    \end{subfigure}
  
    \vspace{0.5em}
  
    \begin{subfigure}[b]{0.48\linewidth}
      \centering
      \includegraphics[width=\linewidth]{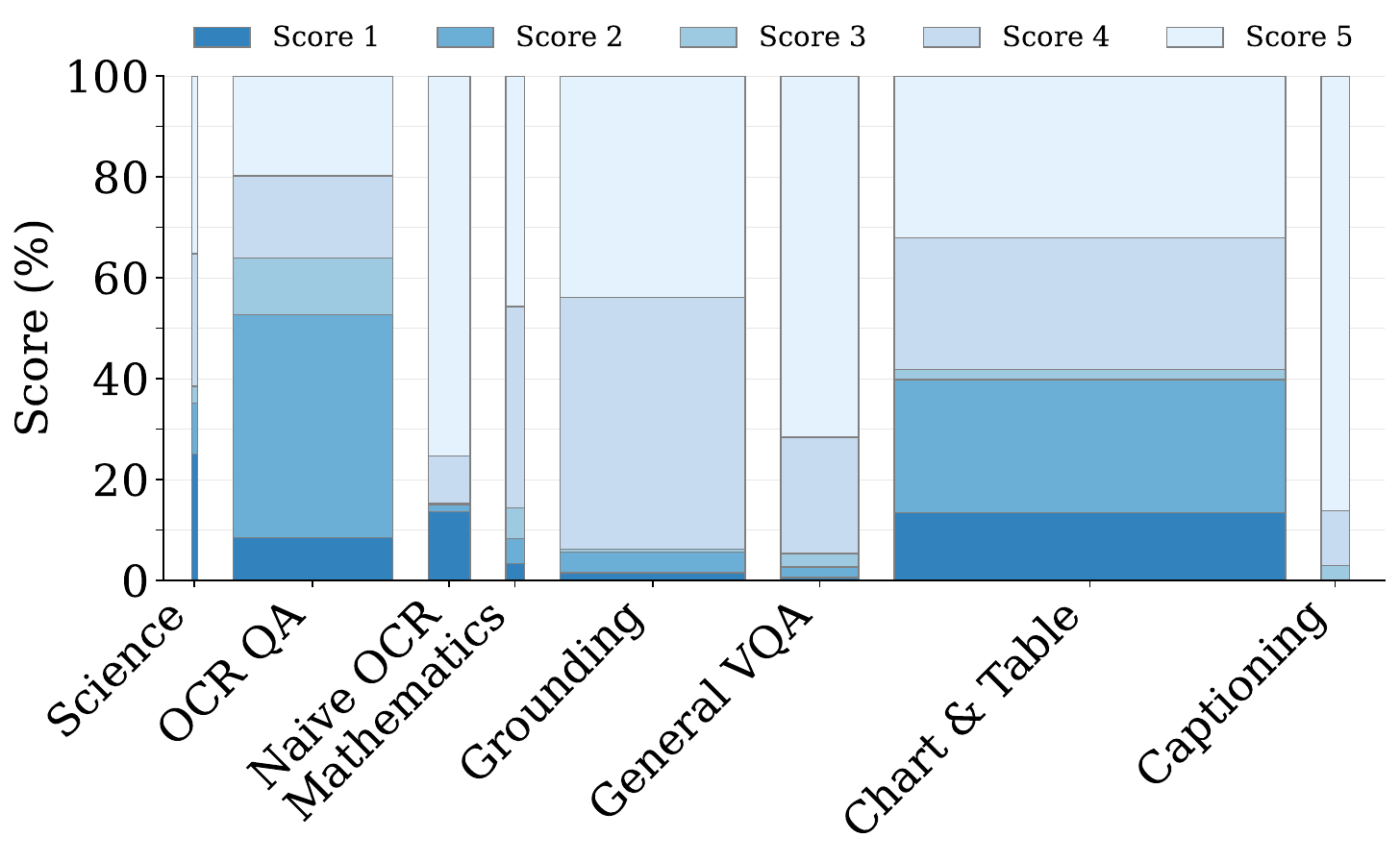}
      \caption{Visual Dependency}
      \label{fig:visual_dependency_scores}
    \end{subfigure}
    \hfill
    \begin{subfigure}[b]{0.48\linewidth}
      \centering
      \includegraphics[width=\linewidth]{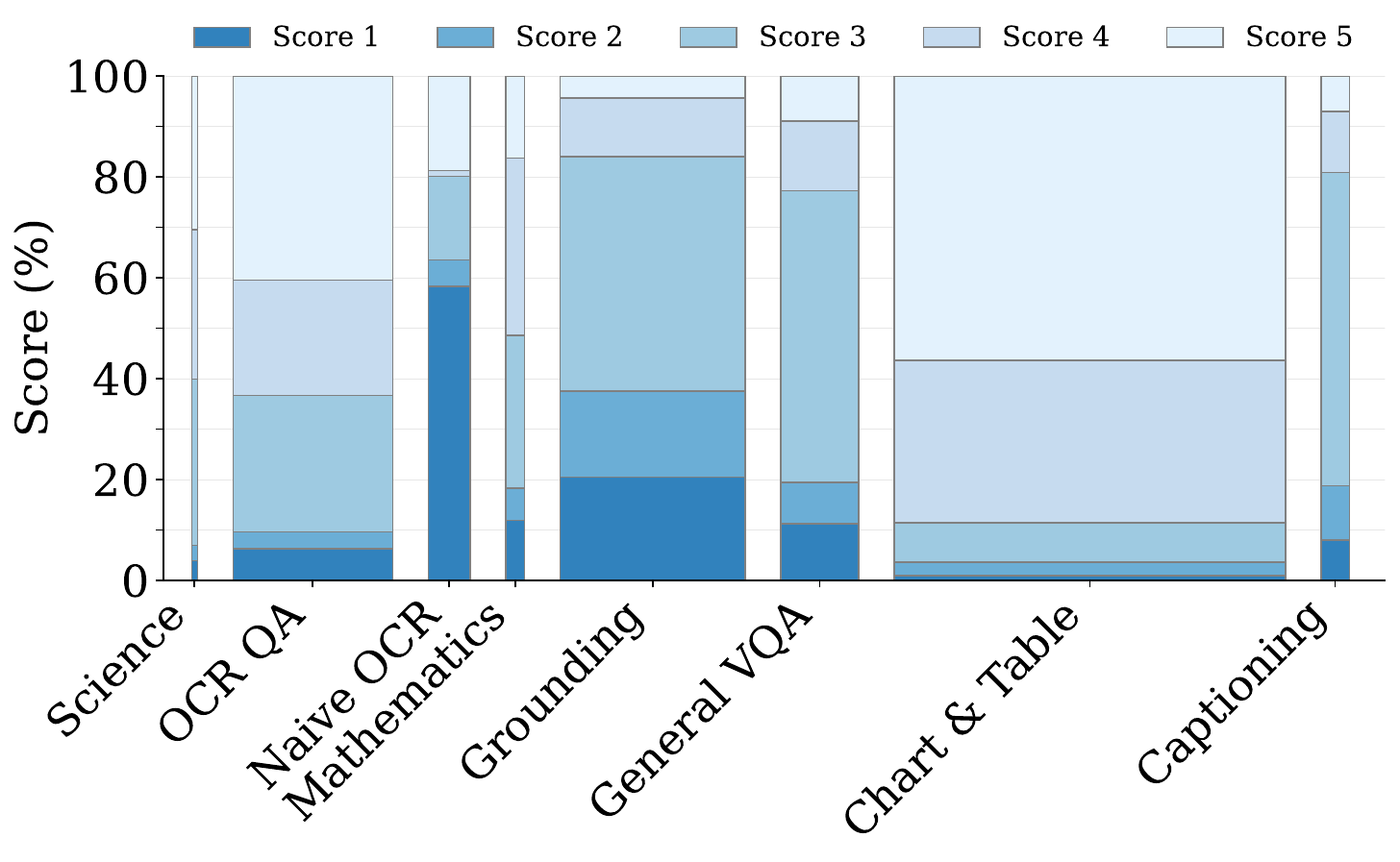}
      \caption{Image Correspondence}
      \label{fig:image_correspondence_scores}
    \end{subfigure}
  
    \caption{\textbf{Quality rating distributions by category.}
  Score distributions across the four quality axes (top left: Formatting, top right: Relevance, bottom left: Visual Dependency, bottom right: Image-Question Correspondence) broken down by dataset category. Category width corresponds to the number of turns.}
  \label{fig:quality_distributions_by_category}
\end{figure}

\begin{figure}
  \centering
  \includegraphics[width=0.8\linewidth]{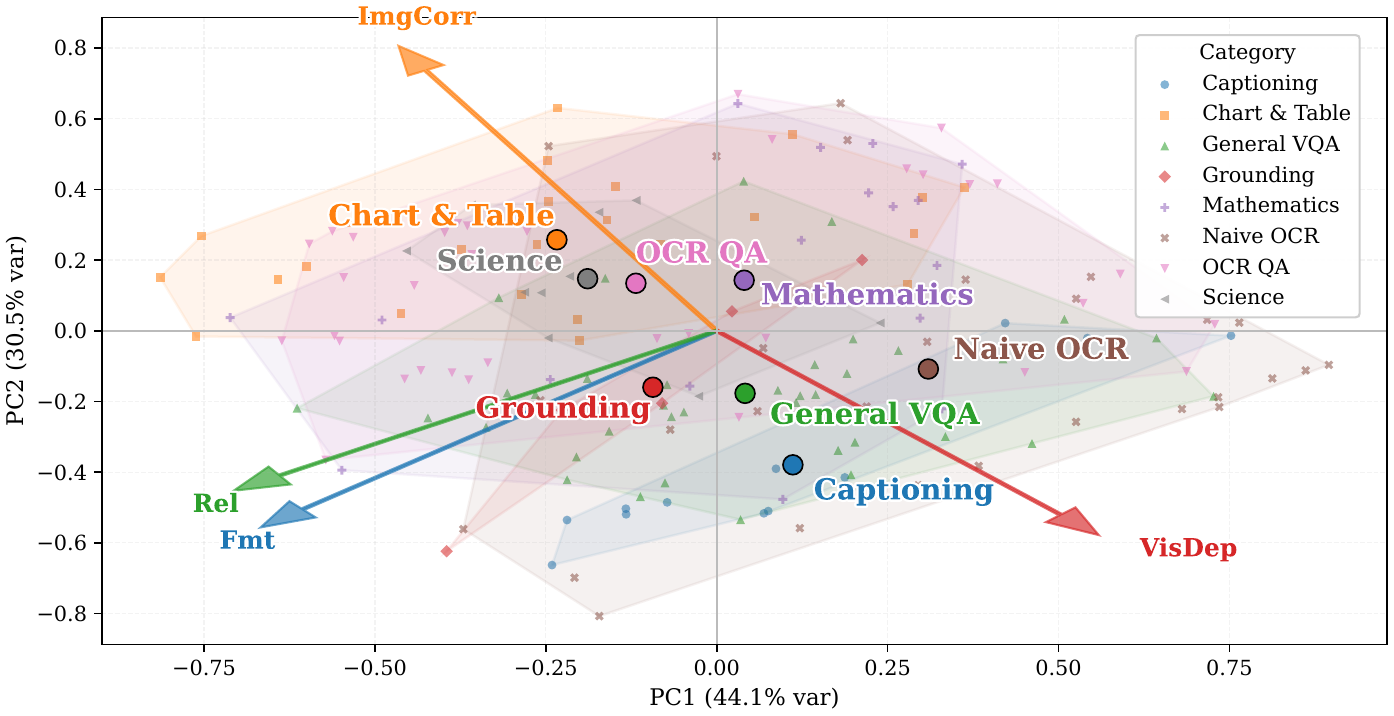}
  \caption{\textbf{Dataset characterization along four axes.} We apply per-dataset PCA over the different characteristic scores. From the analysis, it appears \emph{Formatting} and \emph{Relevance} are highly correlated, while \emph{Visual Dependency} and \emph{Image--Question Correspondence} are strongly inversely correlated. \texttt{Grounding} attains the highest \emph{Formatting/Relevance}, while \texttt{Chart \& Table} attains high \emph{Image--Question Correspondence}. \texttt{Captioning} and \texttt{General VQA} show high \emph{Visual Dependency} combined with strong \emph{Formatting/Relevance}. In contrast, \texttt{Naive OCR} exhibits high \emph{Visual Dependency} but lower scores on \emph{Formatting/Relevance}. Arrows indicate variable loadings; points are dataset centroids with covariance ellipses per category.}
\label{fig:quality_distributions}
\end{figure}

\subsection{Analysis of Characteristic Axes}
\label{sec:char-analysis}

We characterize every training turn by scoring it from 1-5 with LLM/VLM-as-a-judge (\texttt{Qwen3-32B} for text-only criteria and \texttt{Qwen2.5VL-32B-Instruct} for image-conditioned criteria, served locally via vLLM) along four \textbf{characteristic axes}: \emph{Formatting}, \emph{Relevance}, \emph{Visual Dependency}, and \emph{Image--Question Correspondence} (see Appendix~\ref{app:quality_ratings} for the full prompts).
Fig.~\ref{fig:quality_distributions_by_category} shows that \emph{Relevance} is uniformly high across categories, with more than 85\% of the turns scoring 4 or 5, and \emph{Formatting} scores are high overall, with 97.2\% of the turns scoring 4 or 5, peaking for \texttt{Grounding}. These two text-based axes confirm that FineVision pairs well-formed questions with answers that stay on-topic.

As can be seen in Fig.~\ref{fig:quality_distributions_by_category}, the vision-centric axes distinguish task nature most clearly. \texttt{Captioning} and \texttt{General VQA} achieve high scores on both \emph{Visual Dependency} and \emph{Formatting/Relevance}, alongside low scores in \emph{Image--Question Correspondence}. \texttt{Naive OCR} also has high \emph{Visual Dependency}, but with lower scores on the other axes. By contrast, \texttt{Mathematics} shows a different profile, exhibiting lower scores across all four axes. \texttt{Chart \& Table} is defined by high \emph{Image--Question Correspondence} and \emph{Formatting/Relevance}, but lower \emph{Visual Dependency}, mixing low-dependency lookups with higher-dependency integrative cases (\textit{e.g.}, comparing trends across multiple series rather than retrieving a single value), consistent with the variability between reading values and reasoning across trends.

The cross-axis patterns further clarify these roles (see Fig.~\ref{fig:quality_distributions}). The two vision axes -- \emph{Visual Dependency} and \emph{Image--Question Correspondence} -- are inversely correlated, indicating that tasks requiring the image for an answer often differ from those where the question directly corresponds to image content. Conversely, \emph{Formatting} and \emph{Relevance} trend together but remain partly orthogonal to the vision-centric axes. This is evident when comparing \texttt{Grounding}, which scores highly on text-based axes, against \texttt{Naive OCR}, which is highly visually dependent but scores lower on \emph{Formatting} and \emph{Relevance}.
We release per-turn scores to support analysis and reweighting; in our experiments, preserving breadth rather than aggressive filtering yields the best downstream generalization (see Appendix~\ref{sec:quality-filtering}).

\begin{figure}[ht]
  \centering
  \includegraphics[width=0.55\linewidth]{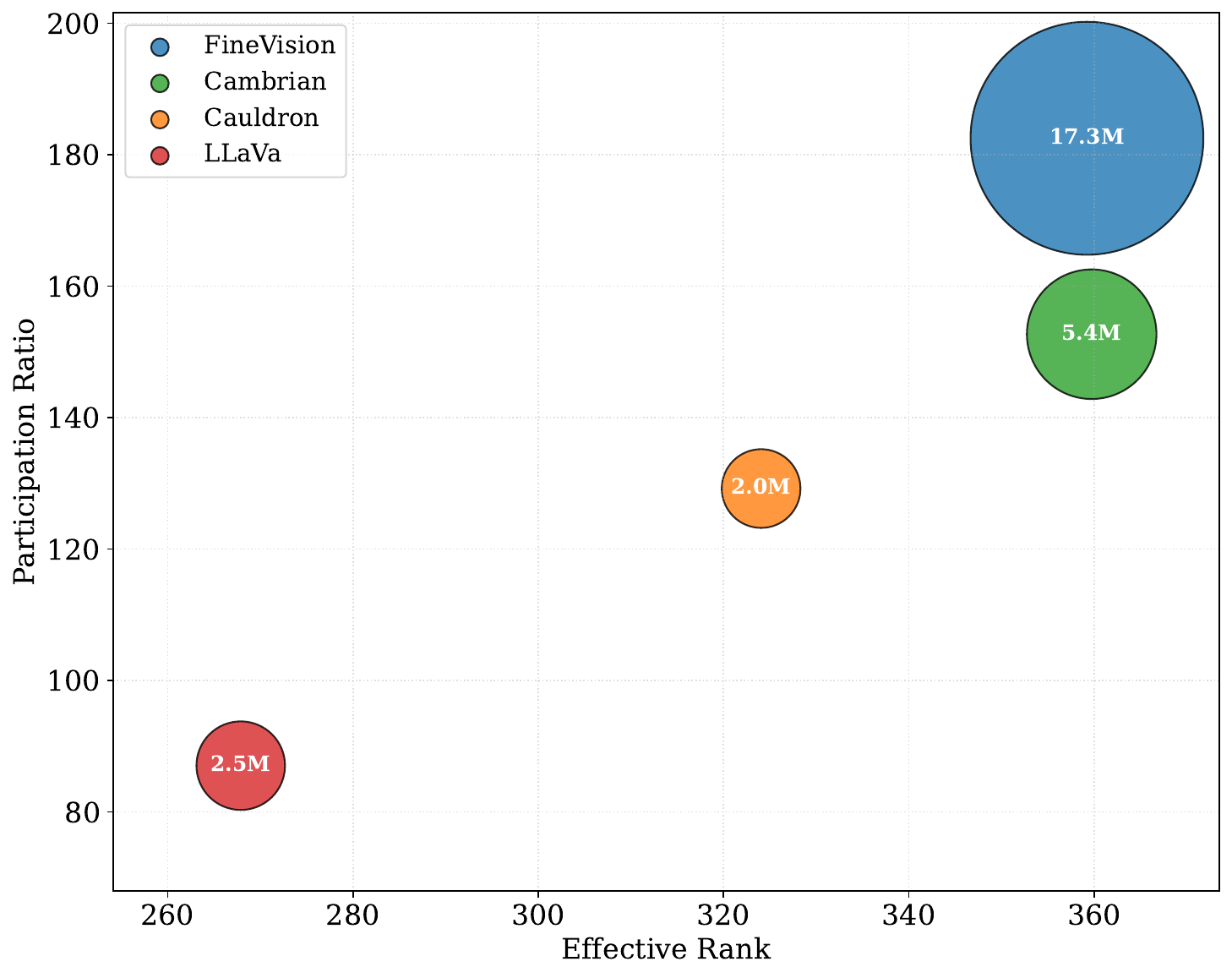}
  \caption{\textbf{Visual diversity analysis.}
While FineVision and Cambrian show similarly high conceptual breadth (effective rank), FineVision exhibits superior conceptual balance (participation ratio).
Higher values are better for both axes; both axes are linear and dimensionless.
Marker size corresponds to the number of images in each dataset. For intuition, a dataset with high effective rank but low participation ratio might cover many animal species but be numerically dominated by a few (\textit{e.g.}, cats/dogs).}
  \label{fig:diversity_composite}
\end{figure}

\subsection{Visual Diversity}
\label{sec:char-diversity}
We analyze visual diversity using covariance-spectrum statistics of self-supervised copy-detection (SSCD) embeddings~\citep{sscd2023} (same pipeline as for deduplication), computed per dataset without subsampling.
SSCD descriptors are optimized to distinguish near-duplicates and, via entropy regularization, promote uniform occupancy of the embedding space, making distances more comparable across regions and suitable for diversity measurement.
Let $\lambda_i$ be the eigenvalues of the embedding covariance; the images are normalized and resized the same way before processing, and the embeddings are not specifically centered.
We report two complementary measures:
\begin{itemize}
    \item \textit{Effective Rank:} $r_{\text{eff}} = \exp(H(p))$ with $p_i = \lambda_i / \sum_j \lambda_j$ and $H(p) = -\sum_i p_i \log p_i$ (equivalent to the Vendi Score~\citep{friedman2023vendi}).
Higher values indicate that the variance is spread across more dimensions, signifying a greater conceptual breadth.
    \item \textit{Participation Ratio:} $\text{PR} = (\sum_i \lambda_i)^2 / \sum_i \lambda_i^2$.
Higher values indicate that the variance is distributed more uniformly across dimensions, indicating a more balanced dataset.
\end{itemize}
As shown in Fig.~\ref{fig:diversity_composite}, these metrics reveal a clear separation of the datasets.
FineVision and Cambrian occupy a high-diversity tier, demonstrating significantly greater conceptual breadth (effective rank) than Cauldron and LLaVA, whose narrower scope may limit the world knowledge of models trained on them.

However, the most crucial insight emerges from the high-diversity tier.
Although both FineVision and Cambrian exhibit a similarly high effective rank, indicating they cover a comparably broad range of visual concepts, FineVision possesses a substantially higher participation ratio.
This distinction is key, as it shows that FineVision's conceptual coverage is not only broad but also significantly more uniform.
Its variance is more evenly distributed between concepts, providing a stronger foundation for training models that are robust and generalize well.
We compute these metrics on the full datasets without subsampling and therefore do not report confidence intervals; given large size differences, naive bootstrapping would be misleading. Covariances are computed in a numerically stable way (\textit{e.g.}, via Welford's algorithm).

Finally, dataset size (marker size) alone does not explain diversity; curation strategy is equally critical.
FineVision's success lies in achieving both massive scale and best-in-class conceptual balance. Exact dataset sizes are reported in Table~\ref{tab:dataset_comparison}.

\section{Experiments and Results}
\label{sec:results}

We conducted a series of experiments to validate the effectiveness of FineVision. We establish the experimental setup, then present our main results comparing FineVision to existing datasets, and finally evaluate novel capabilities.

\subsection{Experimental Setup}

\paragraph{Model and training.}
For all experiments, we train a 460M-parameter SmolVLM~\citep{smolvlm} using the nanoVLM framework \citep{wiedmann2025nanovlm}. 
The architecture consists of a \texttt{SmolLM2-360M-Instruct}~\citep{allal2025smollm2smolgoesbig} text backbone and a \texttt{SigLIP2-Base-512}~\citep{tschannen2025siglip2multilingualvisionlanguage} vision encoder. Unless otherwise specified, we employ a single-stage training protocol for 20,000 steps with an effective batch size of 512, which takes approximately 20 hours on 32 H100 GPUs. With sequence packing to the max length of 8192, this covers more than one effective epoch over the FineVision dataset.

\paragraph{Baselines.}
We compare FineVision against three prominent open-source datasets: The Cauldron \citep{laurencon2024matters}, LLaVA-OneVision \citep{li2024llavaonevision}, and Cambrian-7M \citep{cambrian1}. Table \ref{tab:dataset_comparison} summarizes their respective size and diversity scores.

\begin{table*}[t]
    \centering
    \resizebox{\columnwidth}{!}{%
    \begin{tabular}{lrrrrrrr}
        \toprule
        & \multicolumn{4}{c}{\textbf{Size}} & \multicolumn{2}{c}{\textbf{Diversity}} & \textbf{Performance}\\
        \cmidrule(lr){2-5} \cmidrule(lr){6-7} \cmidrule(lr){8-8}
        \textbf{Dataset} & Images & Samples & Turns & Ans. Tok. & Eff. Rank & Part. Ratio & Avg. Acc.\\
        \midrule
        The Cauldron~\citep{laurencon2024matters} & 2.0M & 1.8M & 27.8M & 0.3B & 324.05 & 129.22 & 38.9\\
        LLaVA-OneVision~\citep{li2024llavaonevision} & 2.5M & 3.9M & 9.1M & 1.0B & 267.89 & 87.05 & 35.9\\
        Cambrian-7M~\citep{cambrian1} & 5.4M & 7.1M & 12.2M & 0.8B & \textbf{359.73} & 152.70 & 44.6\\
        \textbf{FineVision} & \textbf{17.3M} & \textbf{24.3M} & \textbf{88.9M} & \textbf{9.5B} & 359.22 & \textbf{182.52} & \textbf{49.6}\\
        \bottomrule
    \end{tabular}%
    }
    \caption{\textbf{Comparison of dataset size, diversity and final performance.} Size metrics (images, samples, turns, answer tokens), diversity metrics (effective rank, participation ratio) and final performance (average accuracy over 11 benchmarks). FineVision is significantly larger, more diverse, and outperforms comparable open-source alternatives.}
    \label{tab:dataset_comparison}
\end{table*}

\paragraph{Evaluation.}
We use the \texttt{lmms-eval} framework \citep{zhang2024lmms} to evaluate models on a diverse suite of 11 benchmarks, comprising \texttt{AI2D}, \texttt{ChartQA}, \texttt{DocVQA}, \texttt{InfoVQA}, \texttt{MME}, \texttt{MMMU}, \texttt{ScienceQA}, \texttt{MMStar}, \texttt{OCRBench}, \texttt{TextVQA} and \texttt{SEED-Bench}. 

\begin{figure}[]
    \centering
    \includegraphics[width=\linewidth]{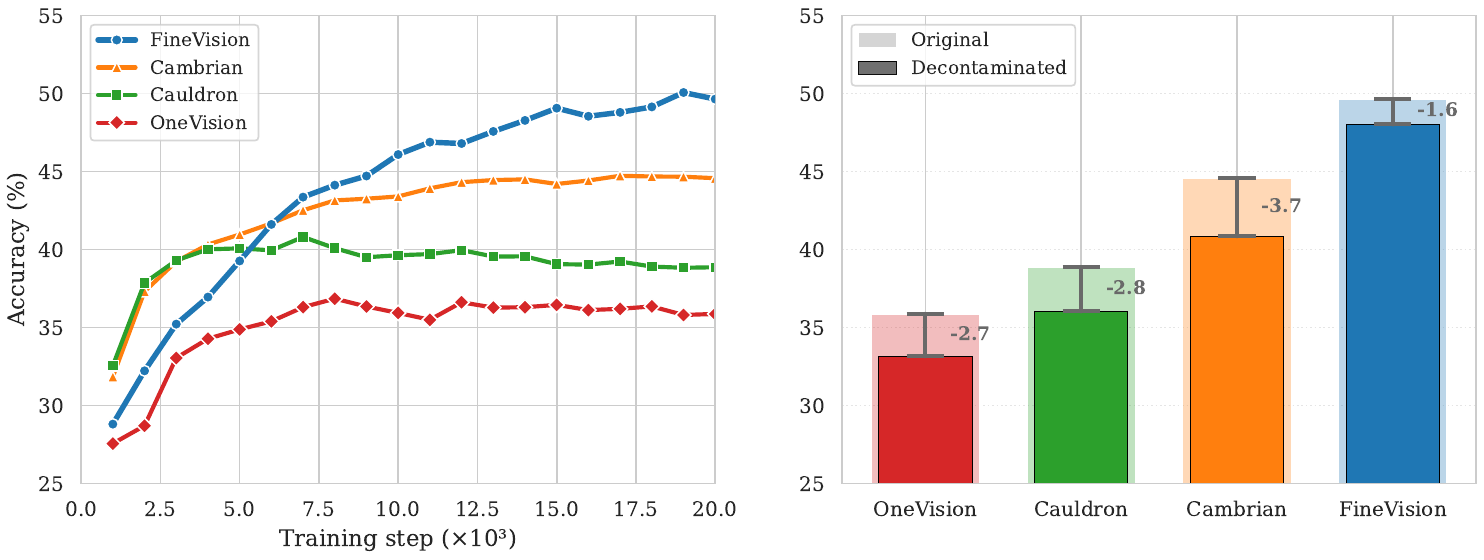}
    \caption{\textbf{Training dynamics on original and decontaminated datasets.} \textbf{Left}: mean normalized performance (\%) across 11 evaluation benchmarks (higher is better), with the training step shown in thousands (\(\times 10^3\)). Each benchmark score is min--max normalized to [0,100] and averaged per evaluation step; the model trained on FineVision (blue) leads throughout the second half of training and attains the best final score. \textbf{Right}: comparison of the final performance between the original data and after decontamination. FineVision exhibits the smallest drop at the end of training with 1.6 pp, whereas baselines degrade by roughly 2.7--3.7 pp, indicating that FineVision's gains are not explained by contamination.}
    \label{fig:against_baselines_combined}
\end{figure}

\subsection{Main Results}
\paragraph{Comparison with existing datasets.}
Models trained on FineVision significantly outperform those trained on other open-source datasets. As shown in Figure \ref{fig:against_baselines_combined}, the FineVision-trained model achieves the highest average performance in all 11 benchmarks (left). Although it initially lags behind during the first few thousand steps -- likely due to the inclusion of novel tasks not present in the baselines -- it surpasses all other models after approximately one epoch of training, demonstrating superior generalization. At the end of training, FineVision produces an average absolute score improvement of 10.8 percentage points (pp) over The Cauldron, 5.1 pp over Cambrian-7M, and 13.8 pp over LLaVA-OneVision.

\paragraph{Impact of test data contamination.}
We investigated test set leakage by processing all datasets through the same pipeline described in Section \ref{sec:dedup} and found that the baseline datasets contain between 2.15-3.05\% of images that are also present in common evaluation benchmarks. 
FineVision has a contamination rate of only 1.02\%. When we retrained all models on decontaminated versions of their respective datasets, the performance of baseline models dropped by 2.7-3.7 pp, while the FineVision model's performance dropped by only 1.6 pp (for full details, see Appendix Table~\ref{tab:contamination_performance}). 

\paragraph{New GUI capabilities.}
FineVision contains substantial amounts of GUI/agentic data, which represents an important new capability for VLMs. 
In addition, tasks measuring performance in this domain are not available in standard evaluation frameworks yet, hindering the widespread tracking of this capability. 
We compare the same 460M model trained on FineVision (\texttt{FV-0.5B}) with an architecturally equivalent SmolVLM2 in two sizes (\texttt{Smol-2B}\footnote{\url{https://huggingface.co/HuggingFaceTB/SmolVLM2-2.2B-Instruct}} and \texttt{Smol-0.5B}\footnote{\url{https://huggingface.co/HuggingFaceTB/SmolVLM2-500M-Video-Instruct}}) on \texttt{Screenspot-V2}~\citep{osatlas2024} and \texttt{Screenspot-Pro}~\citep{li2025screenspot}.
Since these benchmarks are quite challenging for small open models, we report both the performance (accuracy) of the base models as well as after fine-tuning on one epoch of the \texttt{aguvis-stage-1} subset, which is also part of FineVision.
Most small models fail to solve any task at their base stage, and after fine-tuning FineVision-trained models achieve results on par with an architecturally equivalent model 4x its size.

\vspace{1in}
\begin{table}[h]
    \centering
      \begin{tabular}{lrrrrrr}
          \toprule
          & \multicolumn{3}{c}{Base Models} & \multicolumn{3}{c}{FineTuned} \\
          \cmidrule(lr){2-4} \cmidrule(lr){5-7}
          & Smol-2B & Smol-0.5B & FV-0.5B & Smol-2B & Smol-0.5B & FV-0.5B \\
          \midrule
          ScreenSpot-Pro & 0.00 & 0.00 & 0.00 & 0.07 & 0.01 & 0.06 \\
          ScreenSpot-V2 & 0.00 & 0.00 & 0.20 & 0.41 & 0.24 & 0.48 \\
          \bottomrule
      \end{tabular}
      \caption{\textbf{Comparison of model performance on ScreenSpot.} While this benchmark is challenging for small open models, the FineVision-trained model shows strong performance and achieves comparable results to an architecturally equivalent model 4x its size.}
\end{table}

\newpage
\section{Related Work}

\paragraph{Large-scale multimodal data generation pipelines (new data creation).}
This line of work creates new large-scale multimodal datasets via synthetic generation or multi-expert fusion to overcome the scalability limits of human annotation. Early pipelines like \emph{LLaVA-Instruct-150K} ~\citep{llava2023} ($\sim$158K image\textendash instruction pairs over $\sim$118K COCO images) demonstrated GPT-4\textendash generated multimodal instructions guided by BLIP/ CLIP-style embeddings. Specialized generation then scaled in multiple directions: \emph{DenseFusion-1M} ~\citep{li_densefusion-1m_2024} (1.06M pairs from LAION-5B) uses a two-stage perceptual-fusion pipeline that integrates object detectors, OCR, and depth estimators with a multimodal model, including error filtering, to produce hyper-detailed single-paragraph captions; \emph{ShareGPT4V} ~\citep{leonardis_sharegpt4v_2025} develops a seed (100K) to expansion (1.2M) recipe using GPT-4V followed by ShareCaptioner with length/content quality filters; and \emph{WebSight} ~\citep{laurencon_unlocking_2024} synthesizes $\sim$2M webpage screenshots from LLM-generated HTML/CSS (Tailwind), applying rendering/quality filters and removing unsupported/noisy pages to create perfectly aligned UI-image\textendash code pairs. Document-centric pipelines push reading supervision: \emph{DocVLM} ~\citep{nacson2024docvlm} instruments high-resolution documents with OCR for efficient reading, while large meta-collections such as \emph{Docmatix} ~\citep{laurencon2024building} ($\sim$9.5M QA over $\sim$2.4M document images) filter $\sim$15\% hallucinated or unanswerable QAs. Fine-grained generators (\textit{e.g.}, LVIS-\emph{Instruct4V}~\citep{wang2023see}) and region-level prompting (\textit{e.g.}, \emph{ViP-LLaVA-Instruct}~\citep{cai2024vipllava}) emphasize localized grounding, and web-scale interleaved corpora (\textit{e.g.}, \emph{MMC4}~\citep{zhu2023multimodal}, \emph{OBELICS}~\citep{laurenccon2023obelics}) complement instruction data via heavy filtering of raw web documents. Common safeguards across pipelines include expert-fusion signals, rendering/consistency checks, and targeted content/length filters, which together yield denser and more structured supervision than legacy caption/VQA corpora.

\paragraph{Meta-datasets for multimodal instruction tuning.}
The development of large-scale multimodal instruction datasets has rapidly evolved to address the growing demands of vision\textendash language models. 
Early efforts like \emph{MultiInstruct}~\citep{multiinstruct2023} pioneered the field with $\sim$510K fully human-annotated instances across 62 diverse tasks, establishing high-quality instruction-following as a priority. 
\emph{InstructBLIP}~\citep{dai2023instructblip} scaled this approach to $\sim$1.6M instances by aggregating $\sim$12 existing datasets through templated conversion, trading manual curation for breadth. 
The field matured in 2024 with several ambitious collections: \emph{Vision-FLAN}~\citep{xu_vision-flan_2024} brought rigorous human curation to $\sim$1.66M instances across 187 tasks from 101 datasets, emphasizing expert-written instructions; 
\emph{Cambrian-10M}~\citep{cambrian1} pushed scale boundaries with $\sim$10M images and introduced a balanced 7M subset to address quality\textendash quantity trade-offs; 
and \emph{The Cauldron}~\citep{laurencon2024matters} unified 50+ datasets into $\sim$30M dialogue turns for Idefics2, applying targeted test-set decontamination rather than broad internal de-duplication. 
\emph{LLaVA-OneVision}~\citep{li2024llavaonevision} carefully curated $\sim$3.9M instruction\textendash response pairs ($\sim$2.5M images), extending image SFT to multi-image reasoning and video understanding, with strengthened document/OCR and multilingual coverage. 
Most recently, \emph{MAmmoTH-VL-Instruct}~\citep{mammoth2025} demonstrated the potential of fully synthetic pipelines, using open-source models plus filtering to generate $\sim$12M rationale-augmented pairs with detailed reasoning chains, and scaled chain-of-thought supervision for multimodal tasks. 
Other recent works~\citep{eagle2,cho2025perceptionlm} also cite the utilization on the order of 200 datasets. 
Our work, FineVision, addresses these limitations by unifying 200 open sources into a 24M-sample corpus via a semi-automated, human-in-the-loop pipeline that preserves task structure and conversational formatting, applies rigorous intra-/cross-source de-duplication and decontamination against 66 public benchmarks, and extends coverage to agentic/GUI tasks with a unified action space, yielding state-of-the-art results among open-data mixtures.

\paragraph{GUI and embodied vision datasets.}
A newer frontier links perception to action -- models acting in GUIs or embodied environments.
\emph{OS-Atlas}~\citep{osatlas2024} introduced a cross-platform GUI corpus with over 2.3M screenshots and 13M GUI elements spanning web, desktop, and mobile interfaces, and trained a 7B LVLM with a unified function-call API for UI manipulation.
\emph{ShowUI}~\citep{showui2025} presents a vision-language-action model that treats GUI automation as sequence modeling; a lightweight 2B model trained on 256k high-quality interaction steps achieves strong zero-shot grounding.
Complementary efforts target robust GUI grounding and control, including \emph{GUI-Shift}~\citep{gaol2025uishift} and \emph{GUI-Actor}~\citep{microsoft2025guiactor}.
Most GUI agents adopt a unified action space, predicting structured actions (\textit{e.g.}, clicks, typing) as next tokens; cross-platform ambiguities and the limited scale of high-quality interaction data remain open challenges.

\section{Conclusion}
\label{sec:conclusion}
We introduced \textsc{FineVision}, a large-scale, open, and rigorously curated dataset for training vision-language models. Through a semi-automated, human-in-the-loop pipeline that unifies over 200 public sources into a standardized conversational schema, we deliver high-quality supervision spanning captions, VQA, document understanding, OCR, grounding, and GUI interaction. Our pipeline integrates systematic cleaning, near-duplicate control, and benchmark decontamination using SSCD-based matching, enabling reproducible and hygienic training data.

Empirically, models trained on \textsc{FineVision} consistently outperform those trained on existing open datasets across a broad suite of benchmarks, and the gains persist after test-set decontamination. Beyond aggregate scores, \textsc{FineVision} broadens capabilities -- particularly for GUI/agentic settings via a unified action space, suggesting that scale paired with targeted diversity matters for generalization.

We release the dataset, conversion recipes, de-duplication tools, and precomputed embeddings to foster transparent, repeatable research. While our curation reduces leakage and noisy supervision, limitations remain: residual overlaps may persist, long-context and multi-document reasoning are still challenging, and community benchmarks for GUI control are not integrated into the standard evaluation stack. We adhere to source licensing and apply safety-oriented filters; future work will strengthen audits for licensing provenance, privacy, and bias. We view \textsc{FineVision} as a foundation and invite the community to extend it to video, richer multilingual coverage, longer-context reasoning, and stronger human evaluation protocols, further closing the gap between open and proprietary VLM training data.

% Bibliography - using both FineVision and references bib files
\bibliographystyle{hfstyle/plainnat}
\bibliography{references}

\appendix
\newpage
\section{Appendix}
\label{sec:appendix}

\subsection{Duplicate Cluster Visualization}
\label{app:dup_examples}
Visualization of different results from the duplication detection pipeline. Choosing a single threshold to identify duplicated over multiple different categories is a balancing act between false-positives and false-negatives. After manual tuning we settled on $\tau=0.95$.
\begin{figure}[h!]
    \centering
    \begin{subfigure}[t]{0.32\linewidth}
      \centering
      \includegraphics[width=\linewidth]{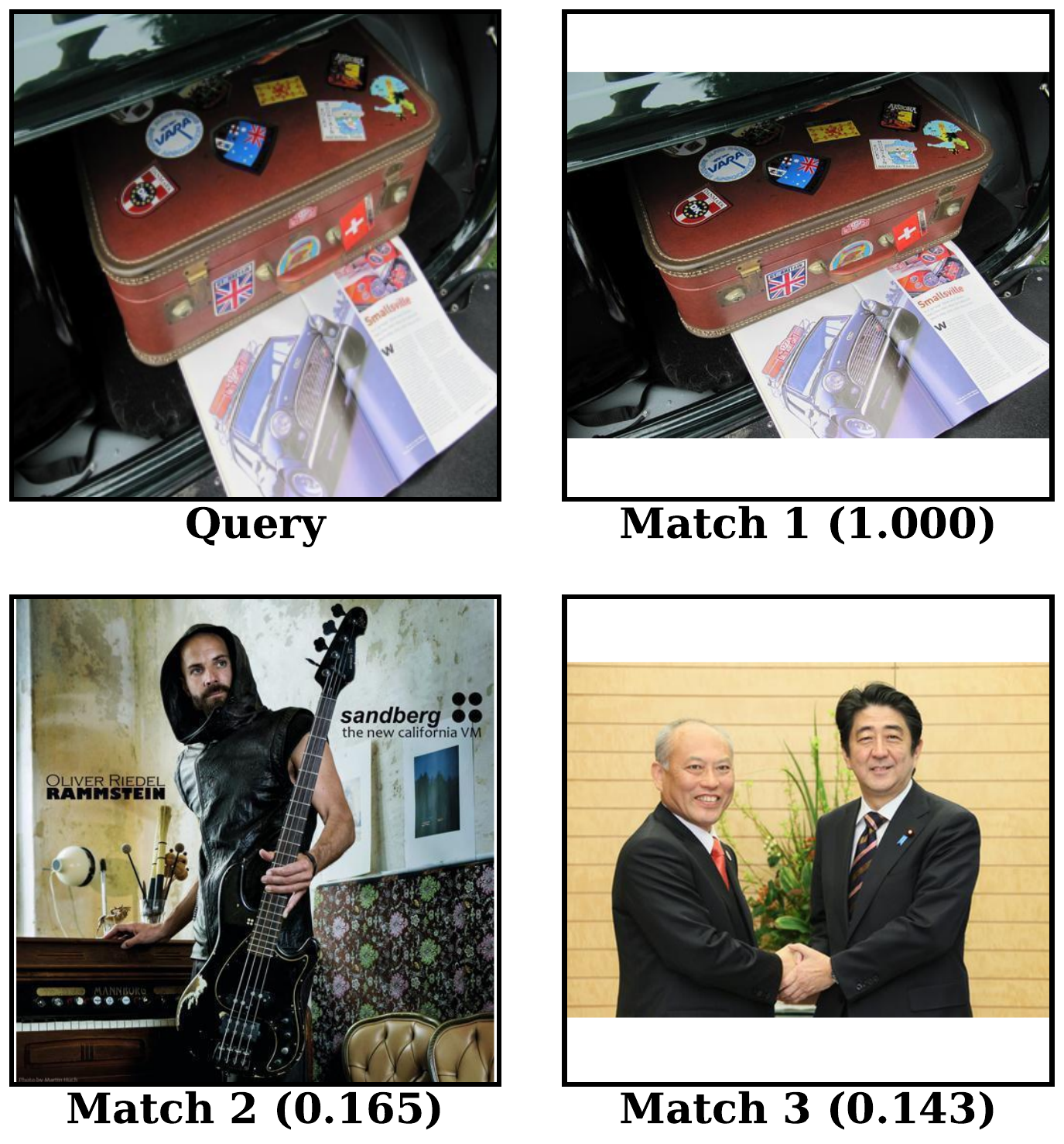}
    \end{subfigure}
    \hfill\vrule\hfill
    \begin{subfigure}[t]{0.32\linewidth}
      \centering
      \includegraphics[width=\linewidth]{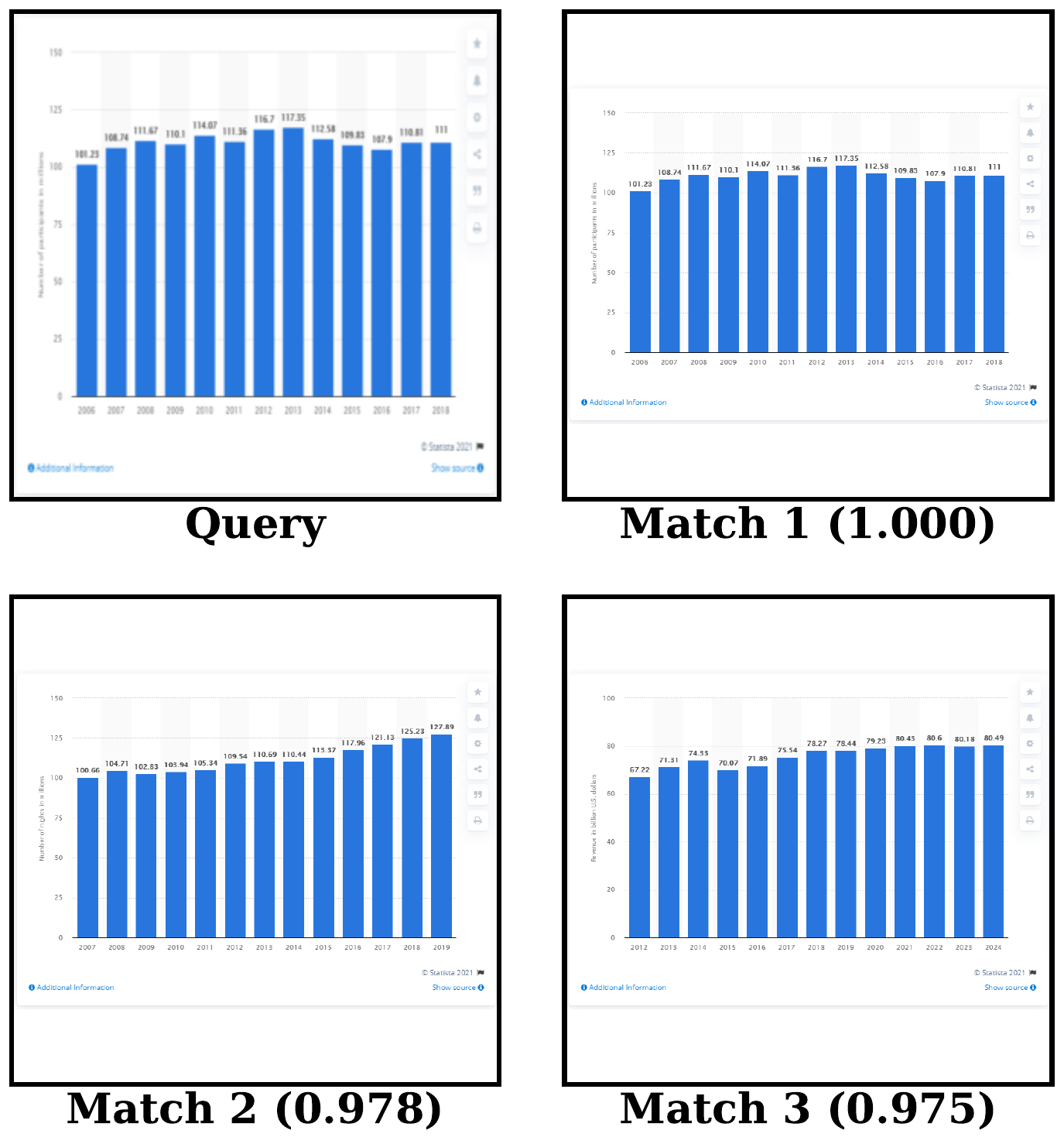}
    \end{subfigure}
    \hfill\vrule\hfill
    \begin{subfigure}[t]{0.32\linewidth}
      \centering
      \includegraphics[width=\linewidth]{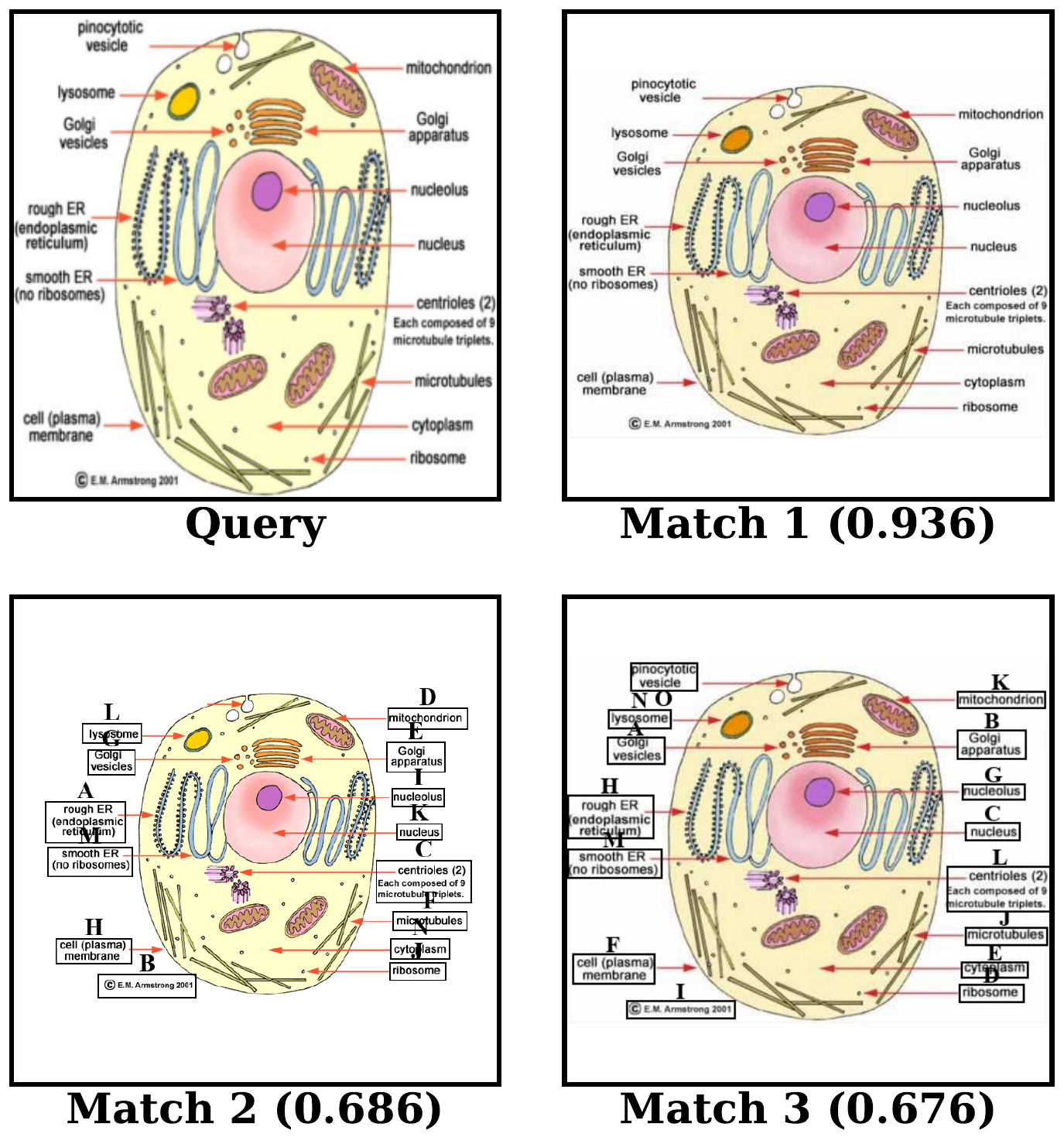}
    \end{subfigure}
    \caption{\textbf{Duplicate detection visualization with $\tau = 0.95$.} Each panel shows the query image and retrieved matches with similarity scores. These three different scenarios, show the difficulty in picking a single threshold: (A, left) true photographic duplicates under mild crops/brightness; (B, middle) false positives (\textit{e.g.}, templated charts with different numbers) just above \(\tau\); (C, right) false negatives (hand drawings) just below \(\tau\). After qualitative experiments we settled on $\tau = 0.95$ since it provided a good trade off between Precision and Recall.}
    \label{fig:dup_examples}
  \end{figure}

\subsection{Quality Ratings}
\label{app:quality_ratings}
These are the full prompts used with the LLM/VLM-as-a-judge pipeline to rate the quality of every turn in FineVision. 

\paragraph{Relevance:}
\begin{verbatim}
Rate how well this answer responds to the question (1-5):

5 - Excellent: Directly and completely answers the question with accurate,
   relevant information
4 - Good: Directly addresses the question with mostly relevant info,
   minor gaps acceptable
3 - Adequate: Partially addresses the question, some relevant information but incomplete
2 - Poor: Minimal attempt to answer, mostly irrelevant or significant gaps
1 - Inadequate: Completely unrelated, only meta-commentary, or unintelligible

RESPOND DIRECTLY WITH ONLY THE NUMBER. NO TEXT, NO EXPLANATION, JUST THE SCORE (1-5).

Question: {question}
Answer: {answer}

Score:    
\end{verbatim}

\paragraph{Formatting:}
\begin{verbatim}
Rate the formatting quality of this text (1-5):

5 - Excellent: Clean, professional, proper grammar/punctuation, well-structured
4 - Good: Generally clean and readable, minor typos that don't impact understanding
3 - Acceptable: Readable despite some formatting issues, occasional special characters
2 - Poor: Significant formatting problems that impact readability, multiple errors
1 - Unacceptable: Severe corruption, extensive encoding errors, or completely garbled

RESPOND DIRECTLY WITH ONLY THE NUMBER. NO TEXT, NO EXPLANATION, JUST THE SCORE (1-5).

Question: {question}
Answer: {answer}

Score:    
\end{verbatim}

\paragraph{Visual Dependency:}
\begin{verbatim}
Rate how much this question depends on visual information to be answered (1-5):

5 - Highly Visual: Requires specific visual details, asks about
   objects/scenes that must be seen
4 - Mostly Visual: Likely requires visual info, asks about visual
   properties or spatial relationships
3 - Moderately Visual: Could benefit from visual info but might be
   answerable with context
2 - Minimally Visual: Primarily answerable from general knowledge,
   visual info provides minor context
1 - Not Visual: Pure general knowledge, abstract concepts, no reference
   to visual elements

RESPOND DIRECTLY WITH ONLY THE NUMBER. NO TEXT, NO EXPLANATION, JUST THE SCORE (1-5).

Question: {question}

Score:   
\end{verbatim}

\paragraph{Image--Question Correspondence:}
\begin{verbatim}
Rate how well this image corresponds to and supports answering the question (1-5):

5 - Perfect: Image directly contains all elements needed, ideal question-image pair
4 - Strong: Image contains most needed elements with clear visual information
3 - Moderate: Image contains some relevant info, partial match with reasonable connection
2 - Weak: Very limited relevant information, mostly unrelated content
1 - No Match: Completely unrelated, corrupted/blank image, or obvious mismatch

RESPOND DIRECTLY WITH ONLY THE NUMBER. NO TEXT, NO EXPLANATION, JUST THE SCORE (1-5).

Question: {question}

Score:   
\end{verbatim}

\newpage
\subsection{Action Space}
\label{app:action_space}
Detailed description of the unified action space.
\begin{table}[h!]
    \centering
    \begin{tabular}{lp{9cm}}
    \hline
    \textbf{Category} & \textbf{Unified Actions} \\
    \hline
    Shared 
        & \texttt{click(x: float, y: float)} \\
        (OS \& Mobile)    & \texttt{type(text: str)} \\
        & \texttt{navigate\_back()} \\
        & \texttt{open\_app(app\_name: str)} \\
        & \texttt{drag(} \\
        & \hspace{1.5em}\texttt{from\_coord: tuple[float, float],} \\
        & \hspace{1.5em}\texttt{to\_coord: tuple[float, float]} \\
        & \texttt{)} \\
    \hline
    OS
        & \texttt{move\_mouse(x: float, y: float)} \\
        & \texttt{double\_click(x: float, y: float)} \\
        & \texttt{right\_click(x: float, y: float)} \\
        & \texttt{press(keys: str | list[str])} \\
        & \texttt{scroll(} \\
        & \hspace{1.5em}\texttt{direction: Literal["up","down","left","right"],} \\
        & \hspace{1.5em}\texttt{amount: int} \\
        & \texttt{)} \\
    \hline
    Mobile
        & \texttt{long\_press(x: float, y: float)} \\
        & \texttt{swipe(} \\
        & \hspace{1.5em}\texttt{from\_coord: tuple[float, float],} \\
        & \hspace{1.5em}\texttt{to\_coord: tuple[float, float]} \\
        & \texttt{)} \\
    \hline
    Completion
        & \texttt{final\_answer(answer: str)} \\
        & \texttt{wait(seconds: int)} \\
    \hline
    \end{tabular}
    \caption{\textbf{Unified action space schema with categories and typed arguments.}}
    \label{tab:unified_action_space}
    \end{table}

\subsection{Data Quality Filtering}
\label{sec:quality-filtering}

\begin{figure*}[]
    \centering
    \begin{subfigure}[b]{0.49\linewidth}
        \centering
        \includegraphics[width=\linewidth]{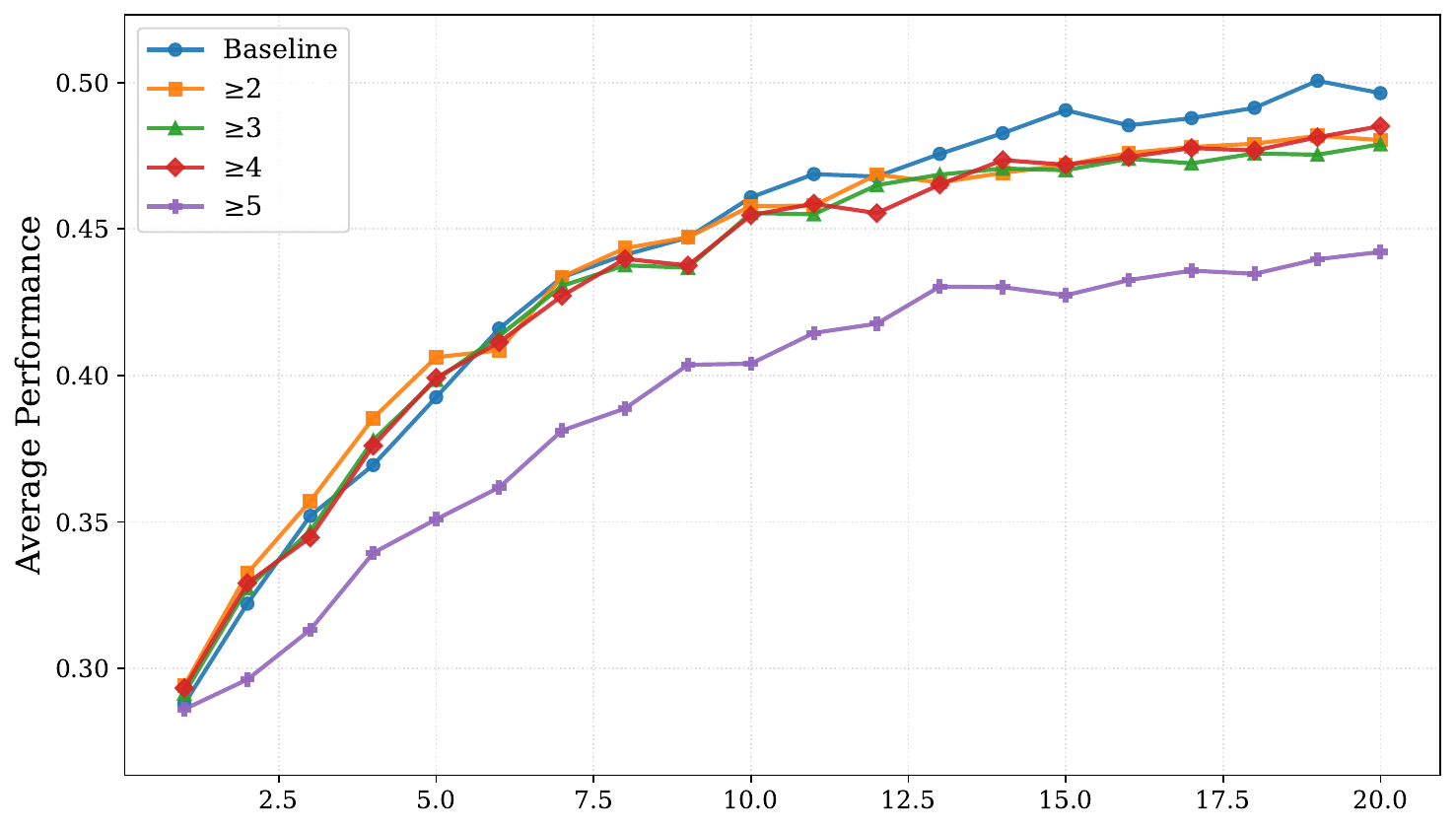}
        \caption{Formatting}
        \label{fig:formatting_filters_rank}
    \end{subfigure}
    \begin{subfigure}[b]{0.49\linewidth}
        \centering
        \includegraphics[width=\linewidth]{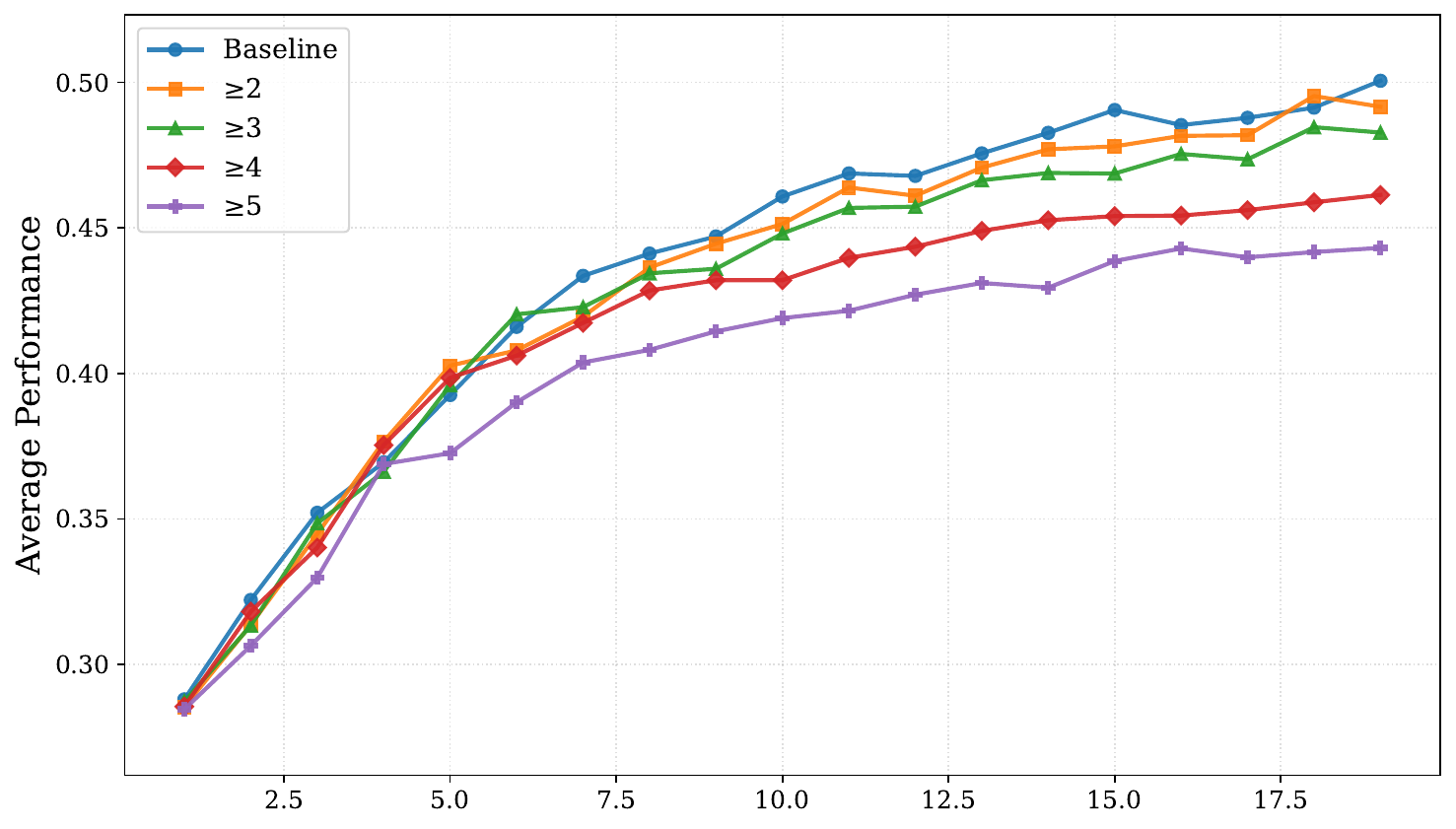}
        \caption{Image Correspondence}
        \label{fig:image_correspondence_filters_rank}
    \end{subfigure}
    \\
    \begin{subfigure}[b]{0.49\linewidth}
        \centering
        \includegraphics[width=\linewidth]{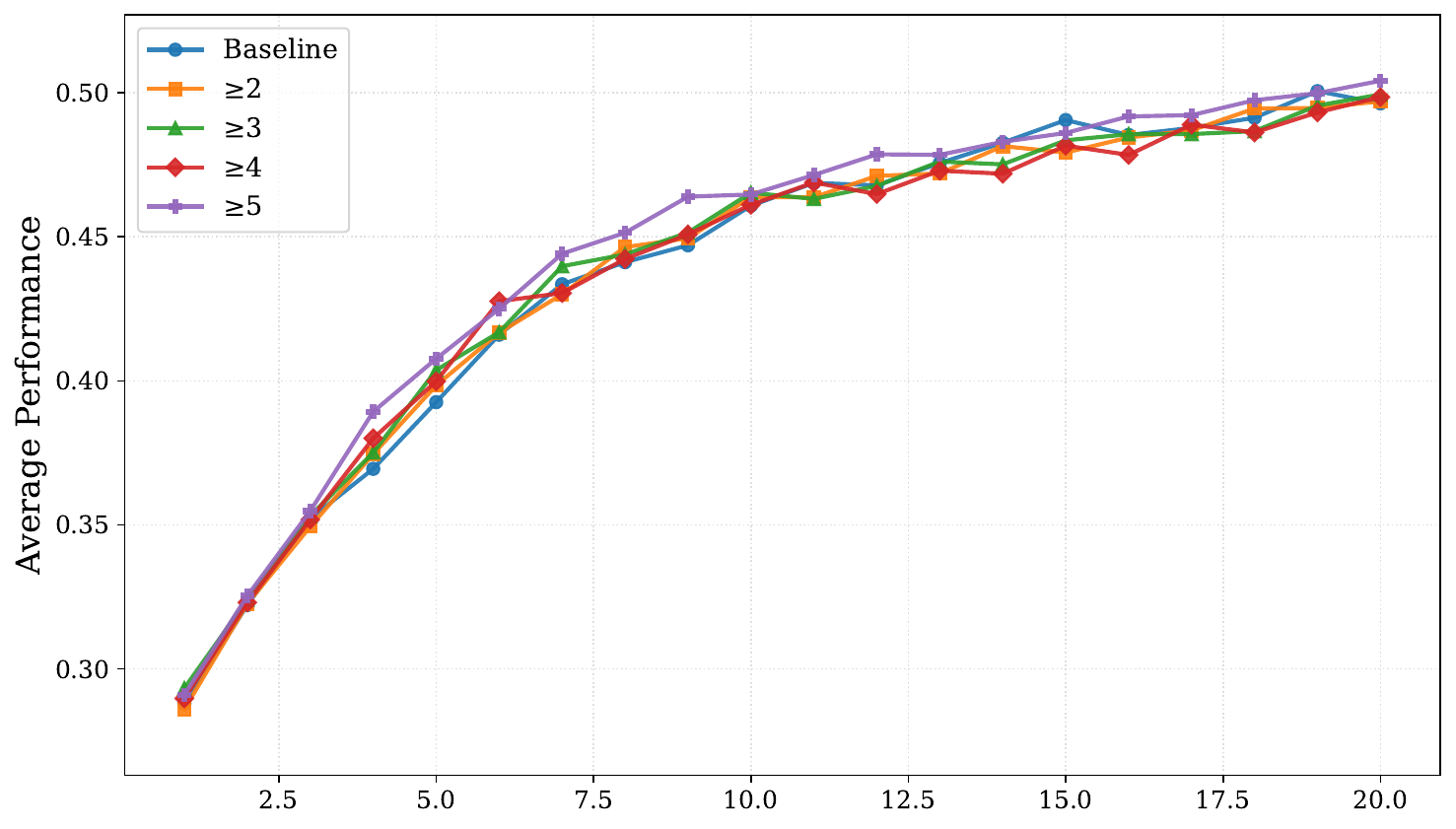}
        \caption{Relevance}
        \label{fig:relevance_filters_rank}
    \end{subfigure}
    \begin{subfigure}[b]{0.49\linewidth}
        \centering
        \includegraphics[width=\linewidth]{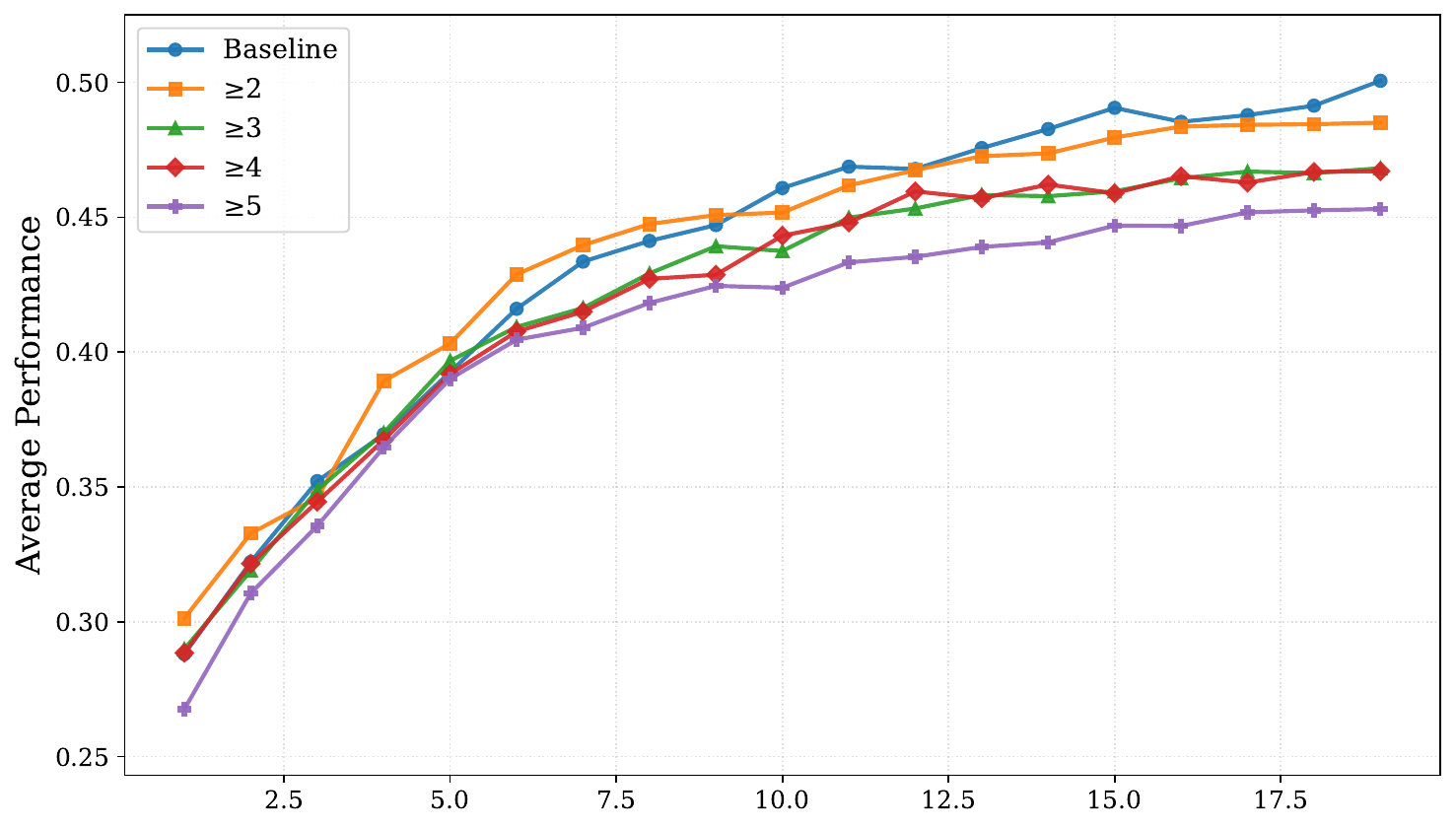}
        \caption{Visual Dependency}
        \label{fig:visual_dependency_filters_rank}
    \end{subfigure}
    \caption{\textbf{Model performance under prompt-based quality filtering.} Average benchmark performance for models trained with thresholds on our four prompt-based quality axes. These results indicate that our specific prompt-based scores are not effective data filters.}
    \label{fig:fl_combined}
\end{figure*}

\begin{figure}[]
    \centering
    \includegraphics[width=0.5\linewidth]{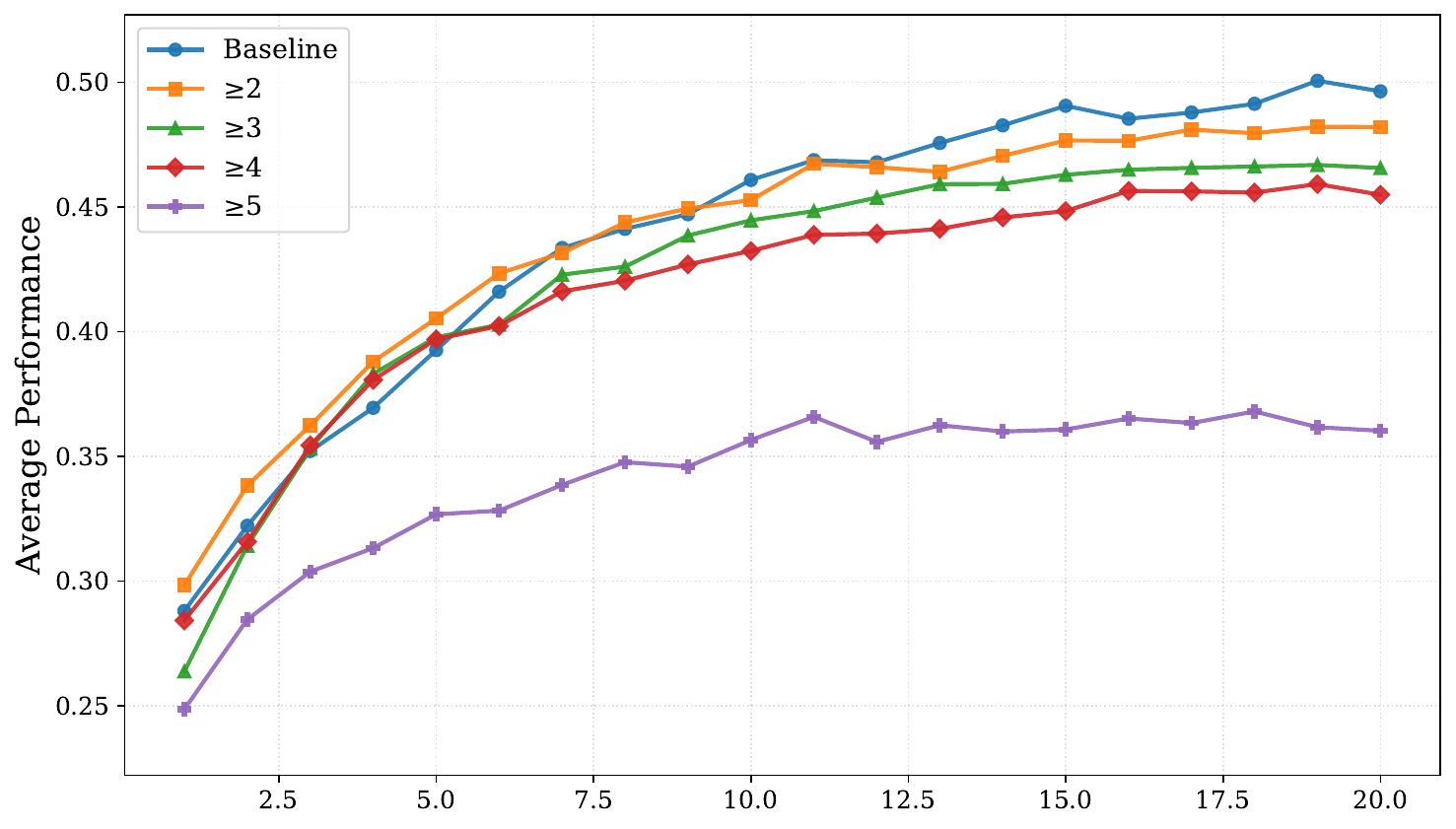}
    \caption{\textbf{Model performance under combined prompt-based quality filtering.} We combine all filters into a single criterion, meaning we only select datapoints that have all four ratings above a certain threshold, training on the full dataset also results in the best performance.}
    \label{fig:all_ratings}
\end{figure}

We evaluated our simple prompt-based quality scores as filters along the four axes defined in Sec.~\ref{sec:char-analysis}.
Across our experiments, these specific scores did not yield an effective filtering scheme: reducing the dataset by thresholding on these metrics generally did not improve model performance compared to training on the unfiltered data (see Fig.~\ref{fig:fl_combined} and \ref{fig:all_ratings}).
This stands in contrast to recent works that report measurable gains from explicit multimodal data selection/cleaning: \emph{Eagle2}~\citep{eagle2} applies rule-based filtering and mixture shaping over large pools; \emph{XMAS}~\citep{xmas} selects via cross-modal agreement trajectories; and \emph{CLEAR}~\citep{chen2024automateddatacurationrobust} retains the most high-quality instructions via confidence-based selection.
Our negative result therefore suggests that how the filter is constructed matters: naive prompt-score thresholding (as instantiated here) is insufficient, whereas targeted procedures (\textit{e.g.}, consistency/difficulty scoring, concept balancing, deduplication) can be beneficial.
Moreover, the goal of filtering itself warrants scrutiny, as some common practices can be actively harmful. For instance, \citet{pouget2024no} demonstrate that filtering web-scale data to English-only pairs degrades a model's cultural understanding and harms performance for underrepresented socioeconomic groups, even while boosting scores on Western-centric benchmarks.
We therefore conclude only that our prompt-based quality metrics, as instantiated here, are not good filters; we do not make claims about other quality estimators or alternative filtering strategies.
To facilitate further work, we release per-turn scores so others can explore different uses or models for data selection.

\subsection{Benchmark Contamination and Effect}
Detailed statistics regarding the benchmark contamination as well as the performance drop after removing these samples.
\begin{table}[h!]
    \centering
    \begin{tabular}{lrrr}
        \toprule
        \textbf{Name} & \textbf{Samples} & \textbf{Contamination Rate} & \textbf{Performance Drop} \\
        \midrule
        Cauldron     & 1.8M  & 3.05\% & 2.8\% \\
        LLaVA-OneVision & 3.9M  & 2.15\% & 2.7\% \\
        Cambrian-7M  & 7.1M  & 2.29\% & 3.7\% \\
        FineVision   & 24.3M & 1.02\% & 1.6\% \\
        \bottomrule
    \end{tabular}
    \caption{\textbf{Contamination and performance drop across datasets.}}
    \label{tab:contamination_performance}
\end{table}

\subsection{Additional Statistics: Token Length, Conversation Turns and Image Resolution by Category}
\label{sec:char-len-res}
\begin{figure}[p]
  \centering
  \includegraphics[width=0.48\linewidth]{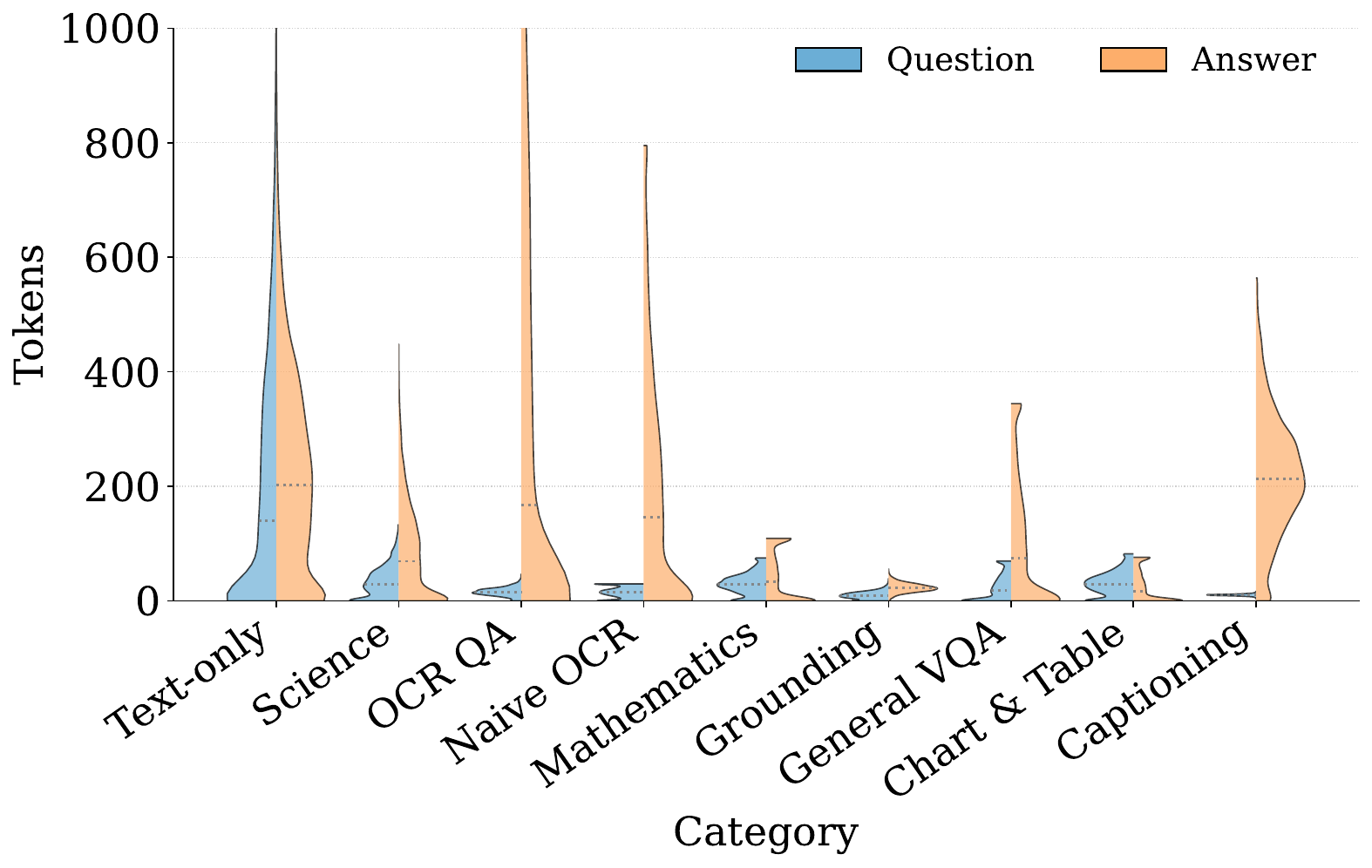}
  \hfill
  \includegraphics[width=0.48\linewidth]{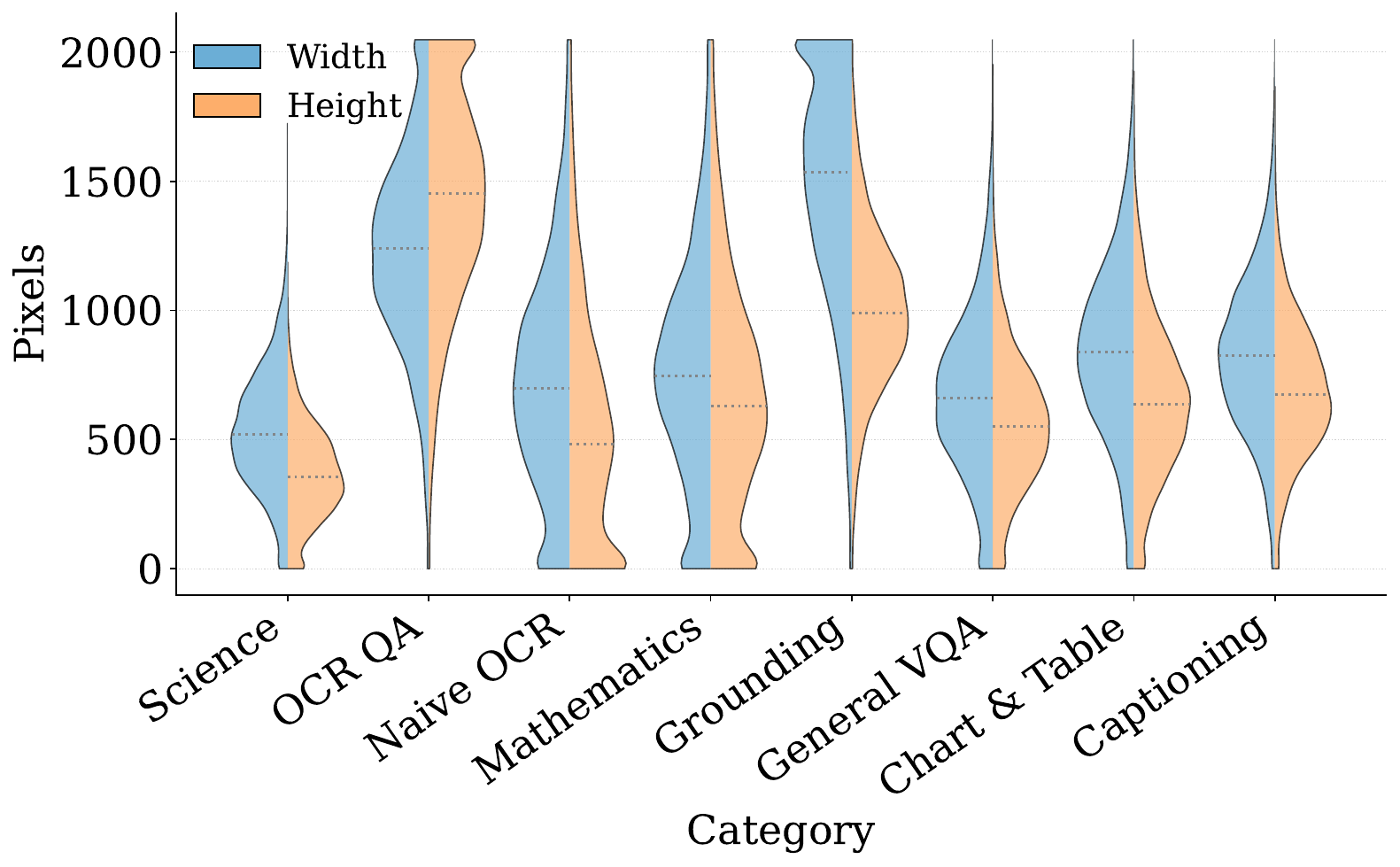}
  \caption{\textbf{Token length and image resolution by category.} \textbf{Left:} split-violin token-length distributions by category; questions (blue) vs. answers (orange), median is dotted line, y-axis capped at 1000 for visibility. These shapes expose task archetypes and the information gap between prompt and response. \textbf{Right:} image resolution distributions by category; width (blue) and height (orange), median is dotted line.}
  \label{fig:length_resolution}
\end{figure}

The split-violin plots in Fig.~\ref{fig:length_resolution} show how interaction type shapes sequence length.
Questions are short and tightly concentrated across categories, whereas answers are broader and often heavy-tailed.
These shapes yield three archetypes: \emph{perceptual/extractive} (\texttt{Grounding}, \texttt{General VQA}, \texttt{Chart \& Table}) with compact distributions; \emph{descriptive generation} (\texttt{Captioning}) with no question and medium-length captions; and \emph{transcription} (\texttt{Naive OCR}), long tail but driven by fidelity rather than inference.
We include both \texttt{Naive OCR} (\textit{e.g.}, ``What is the text in the image?'') and \texttt{OCR QA} (questions that require reading to be answered), treating the latter as more involved reading comprehension and for whole documents.

\begin{figure}[p]
  \centering
  \includegraphics[width=0.6\linewidth]{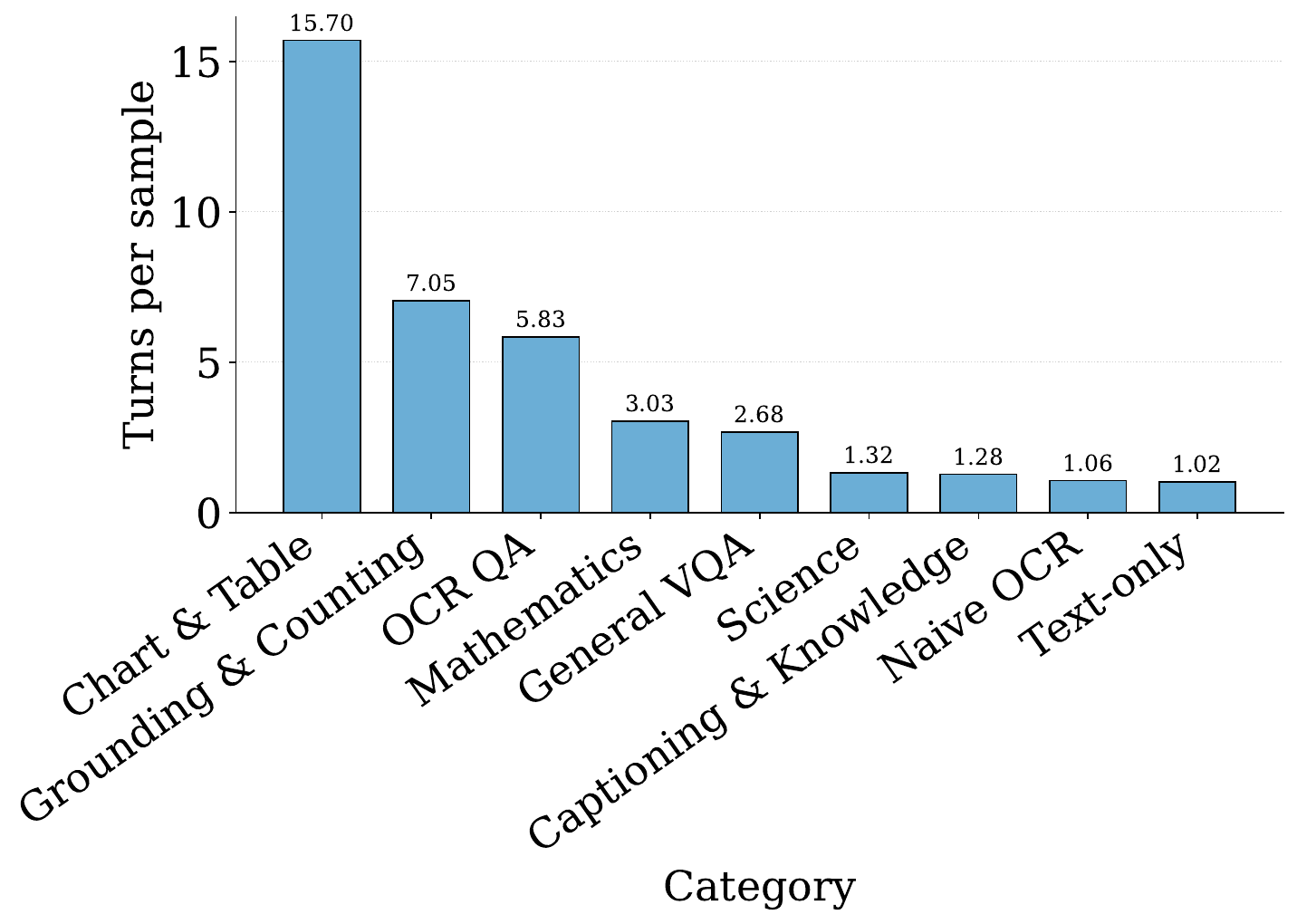}
  \caption{\textbf{Turns per sample by category.} Categories such as \texttt{Chart \& Table} and \texttt{Grounding \& Counting} support more multi-turn interactions per image.}
  \label{fig:turns_per_sample_by_category}
\end{figure}

The disparity between short prompts and longer answers is an information gap the model must fill.
It is largest for \texttt{Naive OCR/Captioning \& Knowledge}, moderate for \texttt{Science}, and minimal for \texttt{Grounding} and \texttt{Chart \& Table}.
Typical median answer-minus-question token gaps by category (dotted lines in Fig.~\ref{fig:length_resolution}) are: Text-only 85.99, Science 33.51, OCR QA 15.33, Naive OCR 104.15, Mathematics 1.36, Grounding \& Counting 12.85, General VQA 45.64, Chart \& Table $-19.26$, and Captioning \& Knowledge 203.47.
\texttt{Chart \& Table} -- and, to a lesser extent, \texttt{Grounding \& Counting / OCR QA} -- naturally support multi-turn exchanges because several queries can target the same figure/document/screenshot.
See also the distribution of turns per sample by category in Fig.~\ref{fig:turns_per_sample_by_category}.
Image resolutions are broadly similar across categories, with document-centric categories (\textit{e.g.} \texttt{OCR QA}) skewing higher to preserve legibility.
Medians are post-resizing and pictured as dotted lines in the figures; median (width, height) by category are: Science (485.03, 332.58), OCR QA (1189.77, 1428.63), Naive OCR (700.56, 443.41), Mathematics (755.31, 608.22), Grounding (1642.53, 950.31), General VQA (641.29, 515.80), Chart \& Table (832.69, 600.97), and Captioning (796.18, 629.51).

\begin{figure}[p]
  \centering
  \includegraphics[width=\linewidth]{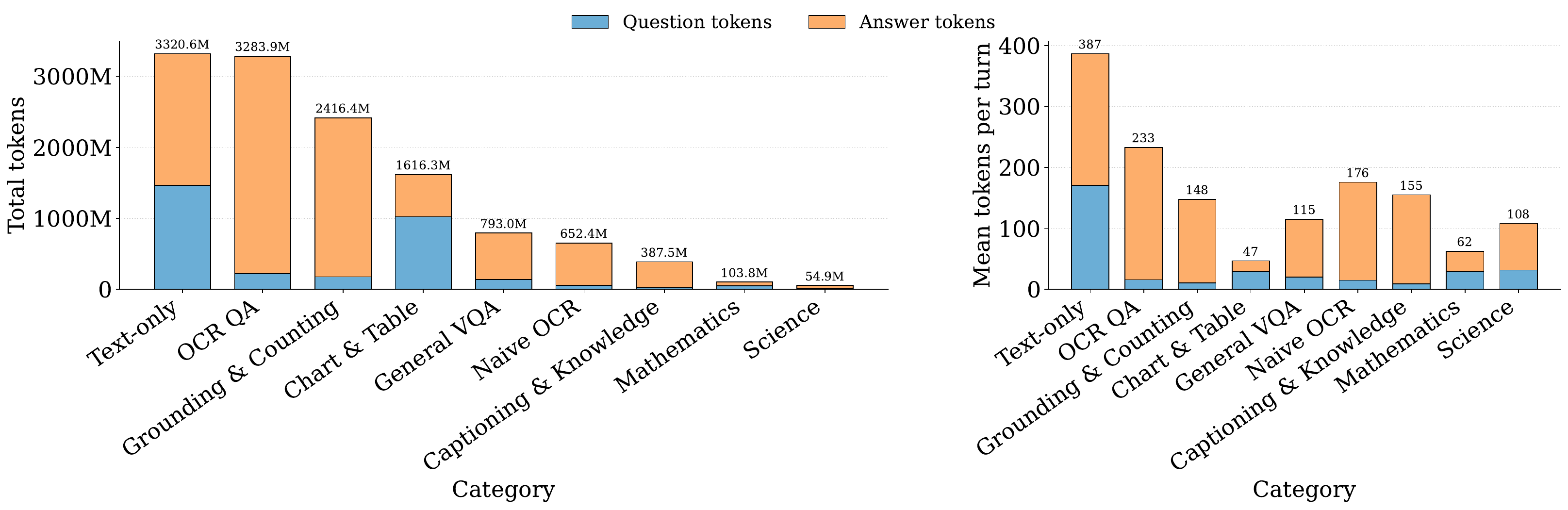}
  \caption{\textbf{Total tokens by category stacked by question and answer.} \textbf{Left:} total tokens, \textbf{Right:} mean tokens per turn}
  \label{fig:tokens_by_category_stacked}
\end{figure}

\newpage
\subsection{FineVision Dataset Subsets}
\label{app:datasets}
Detailed description and statistics of the FineVision dataset subsets by category (see Tables \ref{tab:datasets_captioning_knowledge}, \ref{tab:datasets_grounding_counting}, \ref{tab:datasets_science}, \ref{tab:datasets_mathematics}, \ref{tab:datasets_text_only}, \ref{tab:datasets_chart_table}, \ref{tab:datasets_general_vqa}, \ref{tab:datasets_naive_ocr}, \ref{tab:datasets_ocr_qa}).

% Captioning & Knowledge
\begin{table}[!ht]
    \centering
    \begin{tabular}{lllll}
    \toprule
        Subset Name & Images & Samples & Turns & Answer Tokens \\ \midrule
        coco\_colors~\citep{hazal-karakusmscoco-controlnet-canny-less-colors_nodate} & 118287 & 118287 & 118287 & 6376672 \\ 
        densefusion\_1m~\citep{li_densefusion-1m_2024} & 1058751 & 1058751 & 1058751 & 263718217 \\ 
        face\_emotion~\citep{fast-jobs-face-emotion} & 797 & 797 & 797 & 8066 \\ 
        google\_landmarks~\citep{weyand2020googlelandmarksdatasetv2} & 299993 & 299993 & 842127 & 10202980 \\ 
        image\_textualization(filtered)~\citep{pi_image_2024} & 99573 & 99573 & 99573 & 19374090 \\ 
        laion\_gpt4v~\citep{laiongpt4v-dataset_2023} & 9301 & 9301 & 9301 & 1875283 \\ 
        localized\_narratives~\citep{vedaldi_connecting_2020} & 199998 & 199998 & 199998 & 8021473 \\ 
        sharegpt4o~\citep{cui2025comprehensive} & 57284 & 57284 & 57284 & 36555323 \\ 
        sharegpt4v(coco)~\citep{leonardis_sharegpt4v_2025} & 50017 & 50017 & 50017 & 9825387 \\ 
        sharegpt4v(knowledge)~\citep{leonardis_sharegpt4v_2025} & 1988 & 1988 & 1988 & 293850 \\ 
        sharegpt4v(llava)~\citep{leonardis_sharegpt4v_2025} & 29986 & 29986 & 29986 & 6175899 \\ 
        sharegpt4v(sam)~\citep{leonardis_sharegpt4v_2025} & 8990 & 8990 & 8990 & 1668797 \\ 
        textcaps~\citep{vedaldi_textcaps_2020} & 21906 & 21906 & 21906 & 355991 \\ \bottomrule
    \end{tabular}
    \caption{\textbf{Captioning \& Knowledge datasets.}}
    \label{tab:datasets_captioning_knowledge}
\end{table}

% Grounding & Counting
\begin{table}[!ht]
    \centering
    \begin{tabular}{lllll}
    \toprule
    Subset Name & Images & Samples & Turns & Answer Tokens \\ 
    \midrule
    aguvis-stage-1~\citep{xu_aguvis_2025} & 458957 & 458957 & 3831666 & 93546182 \\ 
    groundui~\citep{zheng_agentstudio_2025} & 13531 & 13531 & 18016 & 883274 \\ 
    objects365\_qa~\citep{shao_objects365_2019} & 1742287 & 1742287 & 12329259 & 2146619635 \\ 
    oodvqa~\citep{tu_how_2023} & 8488 & 8488 & 8488 & 8488 \\ 
    tallyqa~\citep{acharya_tallyqa_2019} & 98680 & 98680 & 183986 & 370282 \\ 
    \bottomrule
    \end{tabular}
    \caption{\textbf{Grounding \& Counting datasets.}}
    \label{tab:datasets_grounding_counting}
\end{table}

% Science
\begin{table}[!ht]
    \centering
    \begin{tabular}{lllll}
    \toprule
    Subset Name & Images & Samples & Turns & Answer Tokens \\ 
    \midrule
    ai2d\_merged~\citep{kembhavi2016diagramworthdozenimages} & 4858 & 4858 & 12325 & 1319140 \\ 
    CoSyn\_400k\_chemical~\citep{yang_scaling_2025} & 8942 & 8942 & 55391 & 2450290 \\ 
    CoSyn\_400k\_circuit~\citep{yang_scaling_2025} & 10470 & 10470 & 67939 & 2637618 \\ 
    pathvqa~\citep{he_pathvqa_2020} & 32632 & 32632 & 32632 & 85168 \\ 
    pmc\_vqa(mathv360k)~\citep{shi_math-llava_2024} & 35948 & 35948 & 35948 & 255109 \\ 
    scienceqa~\citep{lu_learn_2022} & 4976 & 4976 & 6149 & 18447 \\ 
    scienceqa(nona\_context)~\citep{li2024llavaonevision} & 19208 & 19208 & 19208 & 25311 \\ 
    tqa~\citep{kembhavi_are_2017} & 2749 & 2749 & 12567 & 149776 \\ 
    visualwebinstruct(filtered)~\citep{jia_visualwebinstruct_2025} & 263581 & 263581 & 263581 & 31802459 \\ 
    vqarad~\citep{lau_dataset_2018} & 313 & 313 & 1793 & 6003 \\ 
    \bottomrule
    \end{tabular}
    \caption{\textbf{Science datasets.}}
    \label{tab:datasets_science}
\end{table}

% Mathematics
\begin{table}[!ht]
    \centering
    \begin{tabular}{lllll}
    \toprule
    Subset Name & Images & Samples & Turns & Answer Tokens \\ 
    \midrule
    geoqa+(mathv360k)~\citep{cao_augmented_2022} & 17162 & 17162 & 17162 & 117740 \\ 
    unigeo(mathv360k)~\citep{chen2022unigeounifyinggeometrylogical} & 11949 & 11949 & 11949 & 81781 \\ 
    clevr~\citep{lindstrom_clevr-math_2022} & 70000 & 70000 & 699989 & 1570525 \\ 
    clevr\_math~\citep{lindstrom_clevr-math_2022} & 70000 & 70000 & 556082 & 580324 \\ 
    clevr\_math(mathv360k)~\citep{shi_math-llava_2024} & 5280 & 5280 & 5280 & 27536 \\ 
    CoSyn\_400k\_math~\citep{yang_scaling_2025} & 66714 & 66714 & 66714 & 28631388 \\ 
    geo170k(align)~\citep{gao_g-llava_2025} & 35297 & 35297 & 35297 & 1866019 \\ 
    geo170k(qa)~\citep{gao_g-llava_2025} & 12101 & 12101 & 12101 & 1115242 \\ 
    geo3k~\citep{lu_inter-gps_2021} & 2091 & 2091 & 2091 & 2091 \\ 
    geometry3k(mathv360k)~\citep{shi_math-llava_2024} & 9724 & 9724 & 9724 & 69075 \\ 
    geomverse~\citep{kazemi_geomverse_2023} & 9303 & 9303 & 9339 & 2454014 \\ 
    geos(mathv360k)~\citep{seo_solving_2015} & 498 & 498 & 498 & 3509 \\ 
    intergps~\citep{lu_inter-gps_2021} & 1280 & 1280 & 1760 & 5280 \\ 
    mavis\_math\_metagen~\citep{zhang_mavis_2024} & 87348 & 87348 & 87348 & 5486485 \\ 
    mavis\_math\_rule\_geo~\citep{zhang_mavis_2024} & 99986 & 99986 & 99986 & 12535251 \\ 
    raven~\citep{zhang_raven_2019} & 63081 & 42000 & 42000 & 63081 \\ 
    super\_clevr(mathv360k)~\citep{li_super-clevr_2023} & 8642 & 8642 & 8642 & 44129 \\ 
    \bottomrule
    \end{tabular}
    \caption{\textbf{Mathematics datasets.}}
    \label{tab:datasets_mathematics}
\end{table}

% Text-only
\begin{table}[!ht]
    \centering
    \begin{tabular}{lllll}
    \toprule
    Subset Name & Images & Samples & Turns & Answer Tokens \\ 
    \midrule
    text\_ruozhiba~\citep{looksjuicyruozhiba_2024} & 0 & 1496 & 1496 & 234822 \\ 
    text\_code\_feedback~\citep{zheng_opencodeinterpreter_2024} & 0 & 66383 & 221096 & 79752351 \\ 
    text\_codefeedback\_filtered\_instruction~\citep{zheng_opencodeinterpreter_2024} & 0 & 156525 & 156525 & 62764414 \\ 
    text\_infinitymath~\citep{zhang_infinity_2024} & 0 & 101380 & 101380 & 212543 \\ 
    text\_mathinstruct~\citep{yue_mammoth_2024} & 0 & 262039 & 262039 & 44145362 \\ 
    text\_mathqa~\citep{yu2024metamathbootstrapmathematicalquestions} & 0 & 394996 & 394996 & 72451061 \\ 
    text\_mathstepdpo10k~\citep{lai_step-dpo_2024} & 0 & 10795 & 10795 & 989312 \\ 
    text\_numinamath\_cot~\citep{numina_math_datasets} & 0 & 859494 & 859494 & 387758581 \\ 
    text\_openhermes\_2\_5~\citep{OpenHermes_2.5} & 0 & 1001551 & 1008268 & 233561291 \\ 
    text\_openorca~\citep{OpenOrca} & 0 & 4233853 & 4233853 & 468042176 \\ 
    text\_orcamath~\citep{mitra_orca-math_2024} & 0 & 200035 & 200035 & 61860987 \\ 
    text\_pythoncode25k~\citep{flytechpython-codes-25k_2024} & 0 & 49626 & 49626 & 4945892 \\ 
    text\_pythoncodealpaca~\citep{iamtaruncodealpaca_2024} & 0 & 18612 & 18612 & 2683469 \\ 
    text\_theoremqa~\citep{chen_theoremqa_2023} & 0 & 800 & 800 & 3468 \\ 
    text\_wizardlm\_evol~\citep{xu2025wizardlmempoweringlargepretrained} & 0 & 69999 & 69999 & 21955856 \\ 
    text\_OpenMathInstruct-2~\citep{toshniwal_openmathinstruct-2_2024} & 0 & 1000000 & 1000000 & 413132418 \\ 
    \bottomrule
    \end{tabular}
    \caption{\textbf{Text-only datasets.}}
    \label{tab:datasets_text_only}
\end{table}

% Chart & Table
\begin{table}[!ht]
    \centering
    \begin{tabular}{lllll}
    \toprule
    Subset Name & Images & Samples & Turns & Answer Tokens \\ 
    \midrule
    Unichart~\citep{masry_unichart_2023} & 611925 & 611925 & 6898324 & 211989247 \\ 
    tat\_dqa~\citep{zhu_towards_2022} & 2448 & 2207 & 13251 & 1177852 \\ 
    chart2text~\citep{kantharaj_chart--text_2022} & 26961 & 26961 & 30215 & 2670580 \\ 
    chartqa~\citep{masry_chartqa_2022} & 18265 & 18265 & 28287 & 134793 \\ 
    CoSyn\_400k\_chart~\citep{yang_scaling_2025} & 116814 & 116814 & 1085882 & 57641030 \\ 
    CoSyn\_400k\_table~\citep{yang_scaling_2025} & 46518 & 46518 & 416519 & 23335054 \\ 
    dvqa~\citep{kafle_dvqa_2018} & 200000 & 200000 & 2325316 & 5477966 \\ 
    figureqa~\citep{kahou_figureqa_2018} & 100000 & 100000 & 1327368 & 2654736 \\ 
    figureqa(mathv360k)~\citep{shi_math-llava_2024} & 17587 & 17587 & 17587 & 97404 \\ 
    finqa~\citep{chen_finqa_2022} & 5276 & 5276 & 6251 & 224015 \\ 
    hitab~\citep{cheng_hitab_2022} & 2500 & 2500 & 7782 & 335013 \\ 
    lrv\_chart~\citep{li2024llavaonevision} & 1776 & 1776 & 5372 & 158711 \\ 
    mmc\_instruct~\citep{liu_mmc_2024} & 168178 & 168178 & 168178 & 74581055 \\ 
    multihiertt~\citep{zhao_multihiertt_2022} & 30875 & 7619 & 7830 & 244744 \\ 
    plotqa~\citep{methani_plotqa_2020} & 157070 & 157070 & 20249479 & 118122387 \\ 
    robut\_sqa~\citep{zhao_robut_2023} & 8514 & 8514 & 34141 & 1794570 \\ 
    robut\_wikisql~\citep{zhao_robut_2023} & 74989 & 74989 & 86202 & 9276100 \\ 
    robut\_wtq~\citep{zhao_robut_2023} & 38246 & 38246 & 44096 & 6415830 \\ 
    SynthChartNet~\citep{nassar_smoldocling_2025} & 500000 & 500000 & 500000 & 67392223 \\ 
    tabmwp~\citep{lu_dynamic_2023} & 22722 & 22722 & 23021 & 1883243 \\ 
    tabmwp(mathv360k)~\citep{shi_math-llava_2024} & 22452 & 22452 & 22452 & 158042 \\ 
    tat\_qa~\citep{zhu_tat-qa_2021} & 2199 & 2199 & 13215 & 254790 \\ 
    vistext~\citep{tang_vistext_2023} & 9969 & 9969 & 9969 & 1191127 \\ 
    vqaonbd~\citep{vqaonbd_2023} & 39986 & 39986 & 1254165 & 5620523 \\ 
    \bottomrule
    \end{tabular}
    \caption{\textbf{Chart \& Table datasets.}}
    \label{tab:datasets_chart_table}
\end{table}

% General VQA
\begin{table}[!ht]
    \centering
    \begin{tabular}{lllll}
    \toprule
    Subset Name & Images & Samples & Turns & Answer Tokens \\ 
    \midrule
    alfworldgpt~\citep{shridhar_alfworld_2021} & 45073 & 45073 & 45073 & 6276573 \\ 
    chinesememe~\citep{chinese-meme-description-dataset_2024} & 54212 & 54212 & 54212 & 21122723 \\ 
    wildvision~\citep{lu_wildvision_2024} & 333 & 333 & 405 & 72820 \\ 
    allava\_laion~\citep{chen_allava_2024} & 468664 & 468664 & 937328 & 145799426 \\ 
    allava\_vflan~\citep{chen_allava_2024} & 177078 & 177078 & 387872 & 55305642 \\ 
    LLaVA\_Instruct\_150K~\citep{llava2023} & 157710 & 157710 & 361405 & 28719278 \\ 
    datik~\citep{belouadi_automatikz_2024} & 220537 & 222385 & 222385 & 187757952 \\ 
    cambrian(filtered)\_processed~\citep{cambrian1} & 83123 & 83124 & 98534 & 5503211 \\ 
    cocoqa~\citep{ren_exploring_2015} & 46287 & 46287 & 78736 & 212480 \\ 
    CoSyn\_400k\_graphic~\citep{yang_scaling_2025} & 26968 & 26968 & 26968 & 8235679 \\ 
    datikz~\citep{belouadi_automatikz_2024} & 47441 & 47974 & 48296 & 59116193 \\ 
    drivelm~\citep{sima2024drivelm} & 90049 & 4072 & 161030 & 1431417 \\ 
    hateful\_memes~\citep{kiela_hateful_2020} & 8500 & 8500 & 8500 & 17000 \\ 
    iconqa~\citep{lu_iconqa_2022} & 27307 & 27307 & 29841 & 72492 \\ 
    iconqa(mathv360k)~\citep{shi_math-llava_2024} & 22589 & 22589 & 22589 & 134029 \\ 
    idk~\citep{cha_visually_2024} & 11123 & 11123 & 27614 & 665247 \\ 
    indoor\_qa~\citep{keremberke-indoor-scene-classification-2024} & 3350 & 3350 & 3350 & 19700 \\ 
    llavar\_gpt4\_20k~\citep{zhang_llavar_2024} & 19790 & 19790 & 43167 & 1516730 \\ 
    lnqa~\citep{vedaldi_connecting_2020} & 302780 & 302780 & 1520942 & 19027663 \\ 
    lrv\_normal(filtered)~\citep{liu_mitigating_2024} & 10489 & 10489 & 155269 & 3134247 \\ 
    lvis\_instruct4v~\citep{wang2023see} & 222711 & 222711 & 1050622 & 43726782 \\ 
    mimic\_cgd~\citep{laurencon2024matters} & 141878 & 70939 & 141869 & 4304380 \\ 
    mmevol~\citep{luo_mmevol_2024} & 160215 & 160215 & 630441 & 50445237 \\ 
    mmra~\citep{wu_mmra_2024} & 2048 & 1024 & 1024 & 25764 \\ 
    nlvr2~\citep{suhr2017corpus} & 100852 & 50426 & 86373 & 172746 \\ 
    sketchyvqa~\citep{tu_how_2023} & 8000 & 8000 & 8000 & 8000 \\ 
    spark~\citep{yu_spark_2024} & 3904 & 3904 & 6248 & 73973 \\ 
    spatialsense~\citep{yang_spatialsense_2019} & 10440 & 10440 & 17498 & 418883 \\ 
    spot\_the\_diff~\citep{jhamtani_learning_2018} & 17132 & 8566 & 9524 & 209630 \\ 
    vision\_flan(filtered)~\citep{xu_vision-flan_2024} & 175964 & 175964 & 175964 & 3009891 \\ 
    visual7w~\citep{zhu_visual7w_2016} & 14366 & 14366 & 69817 & 209451 \\ 
    vizwiz(mathv360k)~\citep{gurari_vizwiz_2018} & 6604 & 6604 & 6604 & 44876 \\ 
    vqav2~\citep{goyal_making_2017} & 82772 & 82772 & 443757 & 1100837 \\ 
    vsr~\citep{liu_visual_2023-1} & 2157 & 2157 & 3354 & 6708 \\ 
    websight~\citep{laurencon_unlocking_2024} & 10000 & 10000 & 10000 & 5237381 \\ 
    yesbut~\citep{nandy_yesbut_2024} & 4318 & 4318 & 4318 & 157229 \\ 
    \bottomrule
    \end{tabular}
    \caption{\textbf{General VQA datasets.}}
    \label{tab:datasets_general_vqa}
\end{table}

% Naive OCR
\begin{table}[!ht]
    \centering
    \begin{tabular}{lllll}
    \toprule
    Subset Name & Images & Samples & Turns & Answer Tokens \\ 
    \midrule
    ctw~\citep{yuan_large_2019} & 24290 & 24290 & 180621 & 1653254 \\ 
    k12\_printing~\citep{li2024llavaonevision} & 256636 & 256636 & 256636 & 7465001 \\ 
    svrd~\citep{fink_icdar_2023} & 4396 & 4396 & 4396 & 834514 \\ 
    tal\_ocr\_eng~\citep{eagle2} & 256646 & 256646 & 256646 & 7465207 \\ 
    mathwriting-google~\citep{gervais_mathwriting_2025} & 300000 & 300000 & 300000 & 5954806 \\ 
    art~\citep{chng_icdar2019_2019} & 5603 & 5603 & 5603 & 283138 \\ 
    captcha~\citep{captcha2024} & 113062 & 113062 & 113062 & 466856 \\ 
    chrome\_writting~\citep{mouchere_icdar_2013} & 8825 & 8825 & 8825 & 172940 \\ 
    cocotext~\citep{veit_coco-text_2016} & 16169 & 16169 & 16169 & 177111 \\ 
    funsd~\citep{jaume_funsd_2019} & 194 & 194 & 3879 & 29996 \\ 
    hme100k~\citep{yuan_syntax-aware_2022} & 74492 & 74492 & 74492 & 1757743 \\ 
    hw\_squad~\citep{mathew_asking_2021} & 20457 & 20457 & 83682 & 388518 \\ 
    iam~\citep{marti_iam-database_2002} & 5663 & 5663 & 5663 & 130794 \\ 
    iiit5k~\citep{mishra_scene_2012} & 1990 & 1990 & 1990 & 4259 \\ 
    imgur5k~\citep{krishnan_textstylebrush_2023} & 5934 & 5934 & 5934 & 288054 \\ 
    latex\_handwritten~\citep{mouchere_icdar_2013} & 39583 & 39583 & 39583 & 1874733 \\ 
    latexformulas~\citep{oleehyolatex-formulas_2024} & 552340 & 552340 & 552340 & 43094747 \\ 
    maptext~\citep{barney_smith_icdar_2024} & 200 & 200 & 799 & 70813 \\ 
    memotion~\citep{ramamoorthy2022memotion} & 6991 & 6991 & 6991 & 177429 \\ 
    orand\_car\_a~\citep{diem_icfhr_2014} & 1999 & 1999 & 1999 & 9035 \\ 
    rendered\_text~\citep{wendlercrenderedtext_2024} & 10000 & 10000 & 10000 & 244183 \\ 
    sroie~\citep{huang_icdar2019_2019} & 33616 & 33616 & 33616 & 243240 \\ 
    SynthCodeNet~\citep{nassar_smoldocling_2025} & 499983 & 499983 & 499983 & 253422136 \\ 
    synthdog~\citep{avidan_ocr-free_2022} & 500000 & 500000 & 500000 & 48010145 \\ 
    SynthFormulaNet~\citep{nassar_smoldocling_2025} & 499997 & 499997 & 499997 & 51215097 \\ 
    wordart~\citep{avidan_toward_2022} & 19066 & 4804 & 4804 & 54263 \\ 
    olmOCR-mix-0225-documents~\citep{poznanski_olmocr_2025} & 228864 & 228864 & 228858 & 163194337 \\ 
    olmOCR-mix-0225-books~\citep{poznanski_olmocr_2025} & 15194 & 15194 & 15194 & 7962779 \\ 
    \bottomrule
    \end{tabular}
    \caption{\textbf{Naive OCR datasets.}}
    \label{tab:datasets_naive_ocr}
\end{table}

% OCR QA
\begin{table}[!ht]
    \centering
    \begin{tabular}{lllll}
    \toprule
    Subset Name & Images & Samples & Turns & Answer Tokens \\ 
    \midrule
    a\_okvqa~\citep{avidan_-okvqa_2022} & 54602 & 54602 & 54602 & 360990 \\ 
    est\_vqa~\citep{wang_general_2020} & 19358 & 19358 & 19358 & 143270 \\ 
    mmsoc\_memotion~\citep{ramamoorthy2022memotion} & 6991 & 6991 & 6991 & 421250 \\ 
    arxivqa~\citep{li_multimodal_2024} & 100000 & 100000 & 100000 & 6422269 \\ 
    DoclingMatix~\citep{nassar_smoldocling_2025} & 2465202 & 1270911 & 10626898 & 2996338775 \\ 
    ureader\_qa\_processed~\citep{ye_ureader_2023} & 252953 & 252953 & 252953 & 930617 \\ 
    aokvqa~\citep{avidan_-okvqa_2022} & 16539 & 16539 & 17056 & 218917 \\ 
    bentham~\citep{mathew_asking_2021} & 10843 & 10843 & 10843 & 124459 \\ 
    blockdiagramcomputerized~\citep{bhushan_block_2022} & 502 & 502 & 502 & 34453 \\ 
    blockdiagramhandwritten~\citep{bhushan_block_2022} & 1029 & 1029 & 1029 & 75598 \\ 
    CoSyn\_400k\_diagram~\citep{yang_scaling_2025} & 34963 & 34963 & 300357 & 11943321 \\ 
    CoSyn\_400k\_document~\citep{yang_scaling_2025} & 71282 & 71282 & 605173 & 16095526 \\ 
    CoSyn\_400k\_music~\citep{yang_scaling_2025} & 11969 & 11969 & 81786 & 3175586 \\ 
    CoSyn\_400k\_nutrition~\citep{yang_scaling_2025} & 6931 & 6931 & 112097 & 3687254 \\ 
    diagram\_image\_to\_text~\citep{kamizuru00diagram_image_to_text_2024} & 300 & 300 & 300 & 20723 \\ 
    docvqa~\citep{mathew_docvqa_2021} & 10189 & 10189 & 39463 & 275510 \\ 
    handwriting\_forms~\citep{ifthandwriting_forms_2024} & 1400 & 1400 & 1400 & 41490 \\ 
    infographic\_vqa~\citep{mathew_infographicvqa_2022} & 1982 & 4394 & 23717 & 86951 \\ 
    infographic\_vqa\_llava\_format~\citep{mathew_infographicvqa_2022} & 4394 & 2113 & 10054 & 43912 \\ 
    infographic(gpt4v)~\citep{mathew_infographicvqa_2022} & 2113 & 1982 & 1982 & 1044183 \\ 
    invoices\_receipts~\citep{mychen76invoices-and-receipts_ocr_v1_2025} & 3013 & 3013 & 3013 & 771948 \\ 
    mapqa~\citep{chang_mapqa_2022} & 37417 & 37417 & 483416 & 5657339 \\ 
    mapqa(mathv360k)~\citep{shi_math-llava_2024} & 5225 & 5225 & 5225 & 44560 \\ 
    ocrvqa~\citep{mishra_ocr-vqa_2019} & 165746 & 165746 & 801579 & 4801833 \\ 
    pdfvqa~\citep{de_francisci_morales_pdf-vqa_2023} & 8593 & 8593 & 95000 & 939948 \\ 
    screen2words~\citep{wang_screen2words_2021} & 15730 & 15730 & 15743 & 120781 \\ 
    screenqa~\citep{hsiao_screenqa_2025} & 80761 & 80761 & 80761 & 826795 \\ 
    slidevqa~\citep{tanaka_slidevqa_2023} & 11868 & 1919 & 10617 & 156036 \\ 
    st\_vqa~\citep{biten_scene_2019} & 17247 & 17247 & 23121 & 98892 \\ 
    sujet\_finance~\citep{Sujet-Finance-QA-Vision-100k} & 9801 & 9801 & 107050 & 1925361 \\ 
    textocr(gpt4v)~\citep{jimmycartertextocr-gpt4v_2024} & 25060 & 25060 & 25060 & 2436974 \\ 
    textvqa~\citep{singh_towards_2019} & 21953 & 21953 & 34602 & 141882 \\ 
    ureader\_cap~\citep{ye_ureader_2023} & 91215 & 91215 & 91215 & 1435964 \\ 
    ureader\_ie~\citep{ye_ureader_2023} & 17320 & 17320 & 17320 & 128229 \\ 
    ureader\_kg\_processed~\citep{ye_ureader_2023} & 37550 & 37550 & 37550 & 2013731 \\ 
    visualmrc~\citep{tanaka_visualmrc_2021} & 3027 & 3027 & 11988 & 147385 \\ 
    \bottomrule
    \end{tabular}
    \caption{\textbf{OCR QA datasets.}}
    \label{tab:datasets_ocr_qa}
\end{table}

\end{document}